\documentclass[journal,twoside]{IEEEtran}
\IEEEoverridecommandlockouts
\usepackage{cite}
\usepackage{amsmath,amssymb,amsfonts}
\usepackage{algorithm}
\usepackage{algorithmic}
\usepackage{graphicx}
\usepackage{textcomp}
\usepackage{xcolor}
\usepackage{booktabs}
\usepackage{array}

\usepackage{soul}
\usepackage{xcolor}
\usepackage{fancyhdr} % For custom headers/footers

\fancypagestyle{ieee}{ % Custom Header and Footer
    \fancyhf{} 
    \fancyhead[C]{
    \footnotesize This article has been accepted for publication in IEEE Transactions on Circuits and Systems for Artificial Intelligence. This is the author’s version which has not been fully edited and content may change prior to final publication. Citation information: DOI 10.1109/TCASAI.2025.3591073}
    \fancyfoot[C]{
    \footnotesize \copyright~2025 IEEE.  Personal use of this material is permitted.  Permission from IEEE must be obtained for all other uses, in any current or future media, including reprinting/republishing this material for advertising or promotional purposes, creating new collective works, for resale or redistribution to servers or lists, or reuse of any copyrighted component of this work in other works.}

}

\def\BibTeX{{\rm B\kern-.05em{\sc i\kern-.025em b}\kern-.08em
    T\kern-.1667em\lower.7ex\hbox{E}\kern-.125emX}}
\begin{document}

% Flag to manage length of authors' list in references
\bstctlcite{IEEEexample:BSTcontrol}

\title{TrIM, \underline{Tr}iangular \underline{I}nput \underline{M}ovement Systolic Array for Convolutional Neural Networks: \\Dataflow and Analytical Modelling 

\thanks{This work was supported by EPSRC FORTE Programme (Grant No. EP/R024642/2), by EPSRC AI Hub for Productive Research and Innovation in eLectronics (APRIL) (Grant No. EP/Y029763/1), and by the RAEng Chair in Emerging Technologies (Grant No. CiET1819/2/93).}
\thanks{C. Sestito, S. Agwa and T. Prodromakis are with the Centre for 
Electronics Frontiers, Institute for Integrated Micro and Nano Systems,
School of Engineering, The University of Edinburgh, EH9 3BF, Edinburgh, 
United Kingdom. (e-mails: csestito@ed.ac.uk; shady.agwa@ed.ac.uk; 
t.prodromakis@ed.ac.uk).}

}
\author{Cristian~Sestito~\IEEEmembership{Member,~IEEE,} 
        Shady~Agwa~\IEEEmembership{Member,~IEEE,} 
        and~Themis~Prodromakis~\IEEEmembership{Senior Member,~IEEE.}}

\maketitle
\thispagestyle{ieee} % Apply Header and Footer on the first page 
\begin{abstract}
    In order to follow the ever-growing computational complexity and data intensity of state-of-the-art AI models, new computing paradigms are being proposed. These paradigms aim at achieving high energy efficiency by mitigating the Von Neumann bottleneck that relates to the energy cost of moving data between the processing cores and the memory. Convolutional Neural Networks (CNNs) are susceptible to this bottleneck, given the massive data they have to manage. Systolic arrays (SAs) are promising architectures to mitigate data transmission cost, thanks to high data utilization of Processing Elements (PEs). These PEs continuously exchange and process data locally based on specific dataflows (such as weight stationary and row stationary), in turn reducing the number of memory accesses to the main memory. In SAs, convolutions are managed either as matrix multiplications or exploiting the raster-order scan of sliding windows. However, data redundancy is a primary concern affecting area, power, and energy.
    In this paper, we propose TrIM: a novel dataflow for SAs based on a Triangular Input Movement and compatible with CNN computing. TrIM maximizes the local input utilization,  minimizes the weight data movement, and solves the data redundancy problem. Furthermore, TrIM does not incur the significant on-chip memory penalty introduced by the row stationary dataflow.
    When compared to state-of-the-art SA dataflows, the high data utilization offered by TrIM guarantees $\sim 10\times$ less memory access. Furthermore, considering that PEs continuously overlap multiplications and accumulations, TrIM achieves high throughput (up to 81.8\% higher than row stationary), other than requiring a limited number of registers (up to $15.6\times$ fewer registers than row stationary).
\end{abstract}


\section{Introduction}
\IEEEPARstart{N}{owadays}, Artificial Intelligence (AI) is a pervasive paradigm that has changed the way devices can assist everyday activities. However, in order to continuously meet high standards of accuracy, AI models are becoming ever more data-intensive, particularly when Deep Neural Networks (DNNs) are considered. Indeed, other than demanding a huge amount of computations, DNNs also require high memory capacity to manage learned weights, as well as inputs and outputs\cite{8114708}. 

Convolutional Neural Network (CNN)\cite{9451544} is an example of data-intensive DNN, since it executes convolutions on multi-dimensional arrays, named \textit{feature maps}, to carry out tasks like image classification\cite{10.1145/3065386}, image segmentation\cite{9356353,8515234}, image generation\cite{radford2016unsupervisedrepresentationlearningdeep,9892671}, object detection\cite{Redmon_2016_CVPR,Xie_2021_ICCV} and speech recognition\cite{8632885,6857341}. In those applications, CNNs extract local features to determine the intricate relationships of input data. This capability may be further exploited in recent Generative AI models, for example in Convolutional Vision Transformers (CvTs)\cite{Wu_2021_ICCV} and Stable Diffusion models\cite{Zhang_2023_ICCV}. Conventionally, Central Processing Units (CPUs) and Graphics Processing Units (GPUs) manage CNNs' workloads. However, these architectures suffer from the Von Neumann bottleneck\cite{zou_2021}, owing to the physical separation between the computing core and the memory. This significantly degrades the energy efficiency, given that data should be first fetched from an external Dynamic Random Access Memory (DRAM), then buffered on-chip, and finally processed by the computing core. For example, the normalized DRAM energy cost from a commercial 65nm process is 200$\times$ higher than the cost associated to a Multiply-Accumulation (MAC)\cite{10.1109/ISCA.2016.40}, which is the elementary operation performed by the computing core. High data utilization\cite{8097408} is a way to mitigate such cost, by allowing inputs, weights, or partial sums (psums) to be held on-chip as long as they need to be consumed.

Systolic Arrays (SAs) are representative architectures that maximize data utilization by using an array of Processing Elements (PEs) interconnected with each other\cite{10.1145/3604802}. Data moves rhythmically, thus evoking the blood flow into the cardiovascular system, hence the term \textit{systolic}. Conceived late 1970s\cite{kung1982systolic}, they have been mainly used for matrix multiplications. In recent years, the research community has put effort to allow SAs to meet the CNN's workflow\cite{10.1145/3061639.3062207,9238602,10167453}. For instance, the Convolution to General Matrix Multiplication conversion (Conv-to-GeMM)\cite{chetlur2014cudnn} introduces data redundancy to make inputs compatible with SAs' dataflows. However, this reflects in higher memory requirements and, in turn, a higher number of memory accesses, thus negatively impacting area and energy. Weight Stationary (WS) based SAs are examples of architectures using Conv-to-GeMM. Inputs are moved and reused in one direction along the array, while weights are kept stationary. However, First-In-First-Out (FIFO) buffers must assist the data transfer from/to the memory, thus negatively affecting area, power, and energy. The Google Tensor Processing Unit (TPU) is a WS-based SA consisting of $256 \times 256$ PEs, which outperforms CPUs and GPUs by $30\times$ in terms of energy efficiency\cite{10.1145/3079856.3080246}. In the Row Stationary (RS) dataflow\cite{10.1109/ISCA.2016.40}, rows of inputs and weights are reused at the PE level through dedicated memory blocks, without requiring Conv-to-GeMM conversion. In this case, data redundancy is moved at the array level, where inputs are shared diagonally over multiple PEs, while weights are shared horizontally. In addition, inputs and weights circulate cycle-by-cycle inside each PE, thus reducing the energy efficiency, besides making the micro-architecture of PEs more complex. Finally, the area covered by the SA depends on the inputs' sizes, thus making the deployment of large-scale architectures a challenge. Eyeriss\cite{7738524} is the pioneer RS-based SA and consists of 168 PEs, outperforming existing SAs from $1.4\times$ to $2.5\times$ in terms of energy efficiency.

To address the drawbacks of previous art, we propose TrIM, a new dataflow that exploits a triangular input movement to maximize input utilization. A generic SA dealing with TrIM consists of $K \times K$ PEs, where weights are kept stationary and psums are propagated vertically and finally accumulated by an extra adder tree. The TrIM-based SA is compatible with the computational requirements of Convolutional Layers (CLs). 
While we took inspiration from the WS dataflow to keep weights fixed throughout the computations, we considered the convolution-oriented abstraction of the RS dataflow to overcome the Conv-to-GeMM transformation. However, TrIM sits completely apart from the above referenced dataflows because it solves the data redundancy problem: differently from the baseline WS, TrIM does not require Conv-to-GeMM conversion. Also, TrIM is not penalized by the significant energy requirements associated with the on-chip memory blocks unlike the RS dataflow. Furthermore, TrIM avoids the data redundancy at the array level presented by RS.
When compared to WS, TrIM exhibits one order of magnitude less memory accesses. Moreover, TrIM ensures high performance efficiency, as each PE works at the peak throughput of 2 OPs/cycle, which results in 81.8\% improvement of RS. Finally, the micro-architecture of PEs is fairly simple, translating into a limited number of registers, up to $15.6\times$ lower than RS with a kernel size of 7.

The contributions of this work are summarized as follows:
\begin{itemize}
    \item We introduce the TrIM dataflow, which allows high data utilization through a triangular movement of inputs, thus significantly reducing the number of memory accesses from the main memory if compared to previous art.
    \item We present an analytical model for TrIM and state-of-the-art dataflows (i.e., WS and RS), including metrics like memory accesses, latency, throughput, and number of registers. 
    \item Given the analytical model, a thorough design space evaluation is performed, based on several kernel sizes and feature map sizes. WS and RS dataflows are considered for comparisons.
    \item Based on the design space evaluation, we discuss the key benefits offered by TrIM. We also provide a perspective on the exploitation of TrIM in system-level implementations for state-of-the-art AI models.
\end{itemize}

\begin{table}[t]
  \centering
  \renewcommand{\arraystretch}{0.62} % Default value: 1
  \caption{List of Abbreviations}
  \begin{tabular}{c c c c c c}
  \toprule
   Abbreviation & Meaning\\
\bottomrule
\\ AI & Artifical Intelligence \\
\\ CL & Convolutional Layer \\
\\ CNN & Convolutional Neural Network  \\
\\ Conv-to-GeMM & Convolution to GeMM Conversion \\
\\ CPU & Central Processing Unit \\
\\ CvT & Convolutional Vision Transformer \\
\\ DNN & Deep Neural Network \\
\\ DRAM & Dynamic Random Access Memory \\
\\ FCL & Fully-Connected Layer \\
\\ FIFO & First-In-First-Out Buffer \\
\\ fmap & Feature Map \\
\\ FPGA & Field Programmable Gate Array \\
\\ GeMM & General Matrix Multiplication \\
\\ GPU & Graphics Processing Unit \\
\\ $H_I, W_I$ & Height and Width of the input fmap (ifmap) \\
\\ $H_O, W_O$ & Height and Width of the output fmap (ofmap) \\
\\ $I$ & ifmap linear size if square \\
\\ IS & Input Stationary Dataflow \\
\\ $K_H, K_W, K$ & Kernel's Height, Width, Linear size if square \\
\\ L & Latency \\
\\ $M$ & Number of kernels per filters/number of fmaps \\
\\ MA & Memory Access \\
\\ MAC & Multiply-Accumulation \\
\\ $N$ & Number of 3-D filters/number of ofmaps \\
\\ OP & Number of operations \\
\\ OS & Output Stationary Dataflow \\
\\ PE & Processing Element \\
\\ Reg & Number of Registers \\
\\ RS & Row Stationary Dataflow \\
\\ $S$ & Convolutional Stride \\
\\ SA & Systolic Array \\
\\ SRAM & Static Random Access Memory \\
\\ SRB & Shift Register Buffer \\
\\ T & Throughput \\
\\ TPE & Throughput per PE \\
\\ TPU & Tensor Processing Unit \\
\\ TrIM & Triangular Input Movement dataflow \\
\\ WS & Weight Stationary Dataflow \\
\bottomrule
  \end{tabular}
  \label{list_abbrv}
\end{table}
The remainder of this paper is structured as follows: Section II, after introducing CNNs, provides a background about SAs and specific dataflows to manage convolutions; Section III introduces the proposed TrIM dataflow; Section IV presents an analytical model to characterize TrIM, RS and WS; using the referred model, Section V presents an in-depth design space evaluation to spotlight the advantages of TrIM over its counterparts; Section VI provides a perspective on how the TrIM concept can be applied in recent AI models, how can be hardware implemented, and how it compares to the most recent dataflow optimizations; finally, section VII concludes the paper.
In order to facilitate the readability of this manuscript, Table~\ref{list_abbrv} collects all the abbreviations. 

\section{Background}

\subsection{Convolutional Neural Networks}
\begin{figure}
\includegraphics[width=\linewidth]{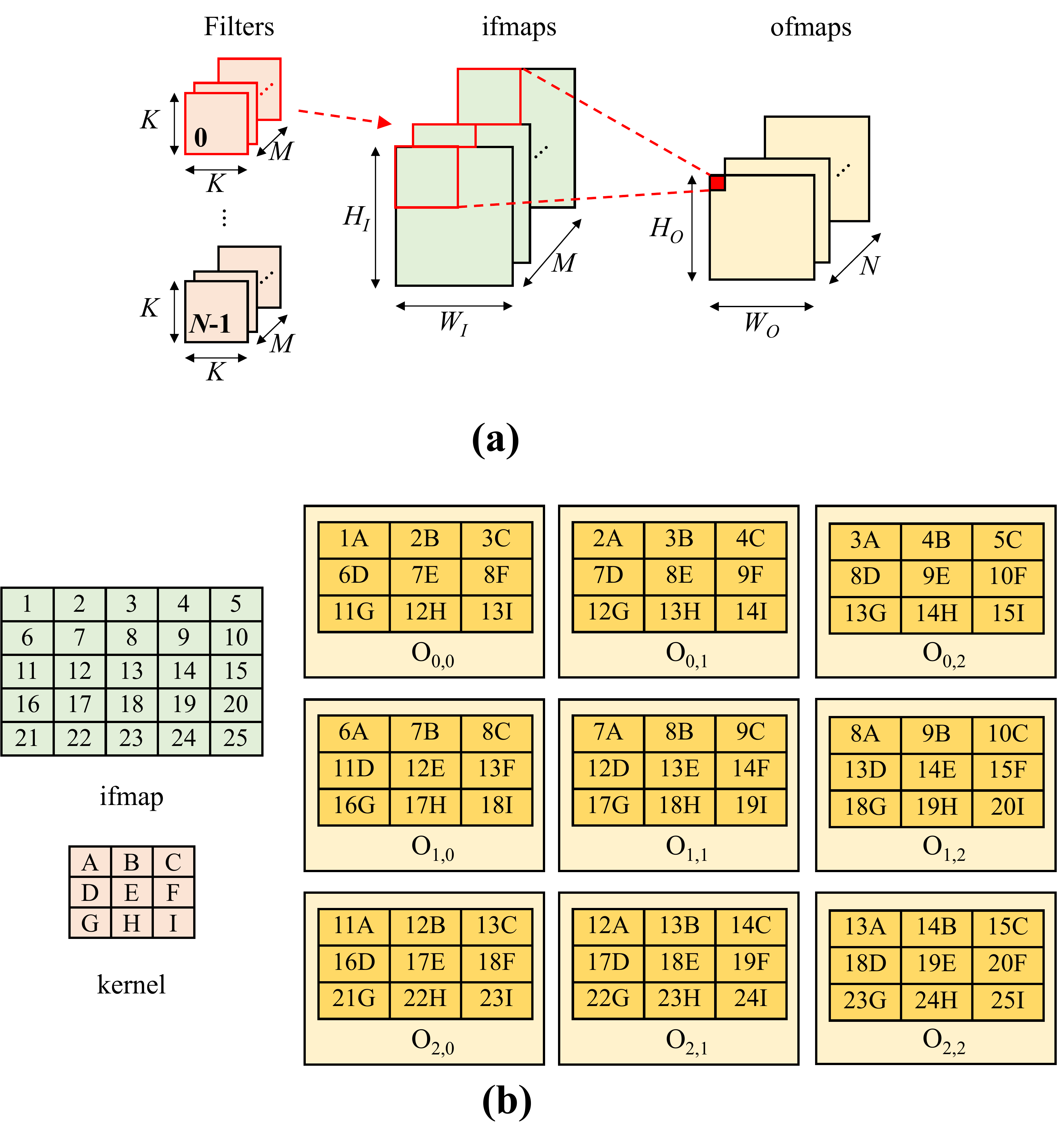}
\centering
\caption{Example of Convolutional Layer (CL): (a) a set of $N$ three-dimensional filters are subjected to as many three-dimensional convolutions with $M$ ifmaps. As a result, $N$ ofmaps are generated. The portion of ifmaps highlighted in red is currently processed by the filters 0, in order to generate the red activation into the first ofmap; (b) example of two-dimensional convolution between a $5 \times 5$ ifmap and a $3 \times 3$ kernel. Each yellow box reports which multiplications are needed to generate the output $O_{i,j}$, with $0 \leq i,j < 3$. 
}
\label{Example_CNN}
\end{figure}

A CNN consists of a sequence of CLs that extract features of interest from input data\cite{lecun2015deep}, by emulating the behavior of the human visual cortex to retrieve patterns, such as shapes and edges, from images. The feature maps (fmaps) generated from the last CL are eventually processed by Fully-Connected Layers (FCLs)\cite{BASHA2020112} for classification. As shown in Fig.~\ref{Example_CNN}(a) Each CL performs $N$ three-dimensional convolutions between $M$ input fmaps (ifmaps), each consisting of a $H_I \times W_I$ plane, and $N$ three-dimensional filters. Each filter volume is made of $M$ matrices that are defined as \textit{kernels}, whose weights are responsible for specific features extraction.
Each filter scans the $M$ ifmaps through sliding windows with stride $S$. In the rest of the paper, we consider $K_H = K_W = K$ and $S = 1$ as in common CNNs for classification. As a result, $N$ output fmaps (ofmaps), each $H_O \times W_O$ wide, are generated, with $H_O = H_I - K + 1$ and $W_O = W_I - K + 1$. Each output element is named activation and follows equation~(\ref{cnn_eq}):
\begin{equation}
\label{cnn_eq}
     O_{n,h_o,w_o} = 
     \sum_{m = 0}^{M - 1} \sum_{k_h = 0}^{K - 1} \sum_{k_w = 0}^{K - 1}
     I_{m, h_o+k_h, w_o+k_w} \times W_{n,m,k_h,k_w}
\end{equation}
$O$ identifies the generic output activation, $I$ is the generic input activation and $W$ is the generic weight belonging to a kernel of one filter; $n$ iterates over the $N$ ofmaps, $h_o$ iterates over the $H_O$ rows of each ofmaps, $w_o$ iterates over the $W_O$ elements belonging to each ofmap's row, $m$ iterates over the $M$ ifmaps, $k_h$ and $k_w$ iterate over the kernels' rows and columns, respectively.
$N$ biases can be eventually added to each output activation if required by the layer. Fig.~\ref{Example_CNN}(b) shows how a convolution between a $5 \times 5$ ifmap and a $3 \times 3$ kernel works.

Finally, CLs may be followed by non-linear functions\cite{7280578} and pooling layers\cite{8648206}, which convert ofmaps in a different numerical domain and downsample their planar sizes, respectively.

\subsection{Systolic Arrays}
A SA is spatial architecture consisting of a 2-D array of PEs interconnected with each other in proper directions. Each PE usually performs a MAC operation using inputs supplied by either the main memory or neighbor PEs. With specific reference to DNNs, PEs are fed by input activations (inputs), weights and psums. High data utilization is met at two hierarchical levels: (a) at the PE level, one element among the input, weight and psum is retained in a register as long as it is required; (b) at the SA level, the other elements move rhythmically between adjacent PEs. The reused data at the PE level equip the SA with a specific stationary dataflow:
\begin{itemize}
    \item Weight Stationary (WS): weights are retained inside the PEs 
    and do not move. Inputs and psum move between adjacent 
    PEs throughout the process.
    \item Input Stationary (IS): a batch of inputs is 
    retained inside the PEs, while weights and psums move between 
    adjacent PEs throughout the process.
    \item Output Stationary (OS): psums are reused at the PE level until 
    the final sum is generated. Inputs and weights move 
    between adjacent PEs throughout the process.
    \item Row Stationary (RS): rows of inputs and 
    weights are stored and reused at the PE level, using memory blocks 
    that enable data circulation. Psums move between adjacent PEs 
    throughout the process.
\end{itemize}
Since CNNs exploit \textit{weight sharing} to process each fmap, WS and RS are commonly used. In the following sub-sections, the two dataflows are presented in detail.

\subsubsection{Weight Stationary SAs}
\begin{figure}
\includegraphics[width=\linewidth]{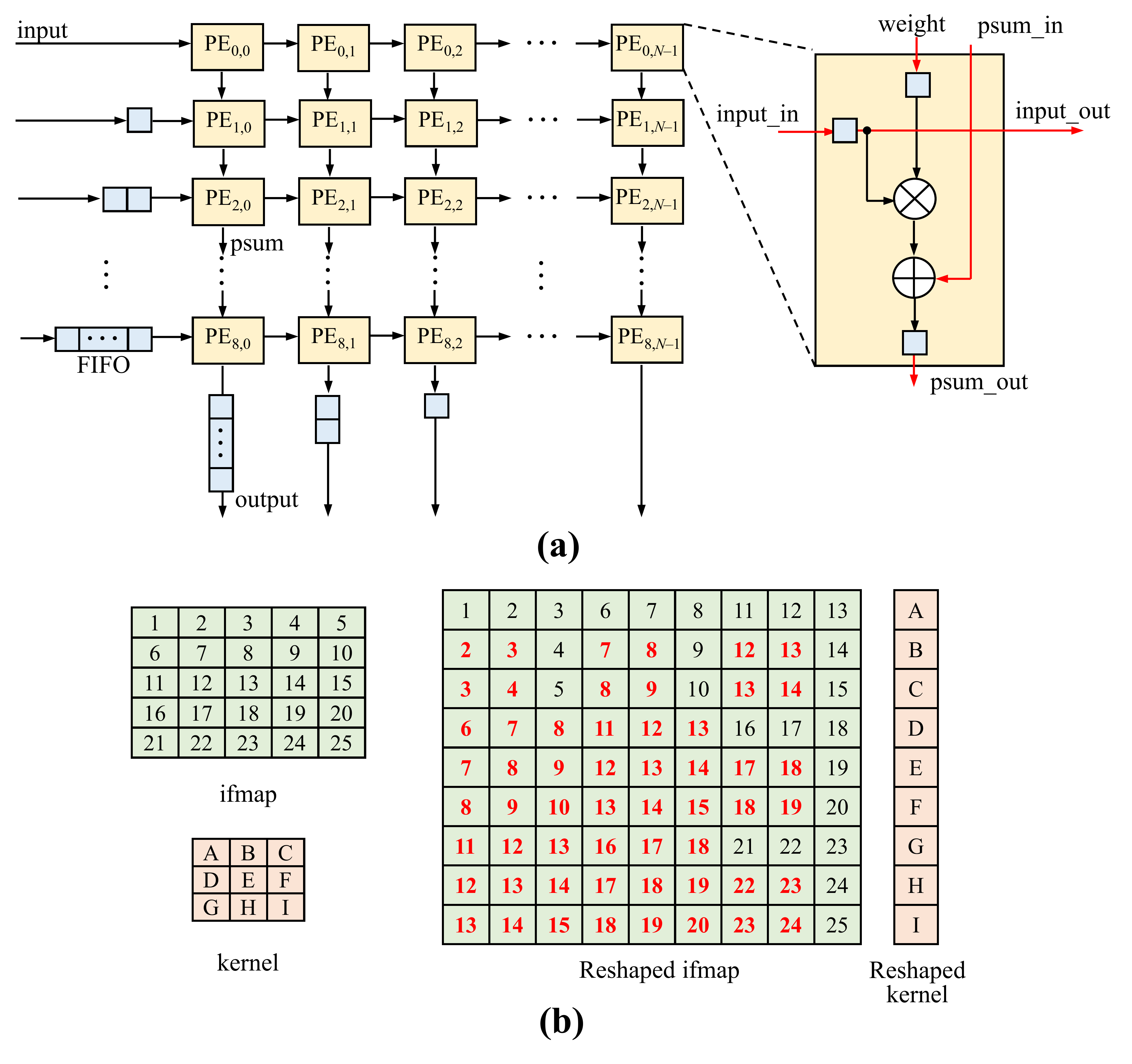}
\centering
\caption{WS-based SA: (a) a general architecture to manage filters with $3 \times 3$ kernels is reported. The different PEs' columns manage the kernels of $N$ different filters, which in turn receive the same ifmap. Inputs move from left to right, while psums move from top to bottom. FIFOs are needed to correctly align inputs and psums over the time. The generic PE performs a MAC operation between the current input, weight and psum as well. Each data is registered to enable proper synchronization; (b) example of General Matrix-Multiplication (GeMM) between a $5 \times 5$ ifmap and a $3 \times 3$ kernel: the ifmap is reshaped as a $9 \times 9$ matrix, where redundant data are highlighted in red, while the kernel is reshaped as a vertical $9 \times 1$ array.}
\label{Example_WS}
\end{figure}
In a WS-based SAs, weights are preliminary fetched from the main memory and stored inside the PEs. Fig.~\ref{Example_WS}(a) depicts an example of WS-based SA, where inputs are loaded from the left side and moved horizontally across the array. Differently, psums are accumulated following vertical interconnections. In order to correctly align data over time, First-In-First-Out buffers (FIFOs) for inputs and psums are placed at the left and bottom boundaries of the array, respectively. To make WS-based SAs compatible with CNNs, inputs and weights are subjected to Conv-to-GeMM\cite{chetlur2014cudnn}. The $M$ ifmaps, consisting of $H_I \times W_I$ elements each, are reshaped as a matrix having $H_O \times W_O$ rows and $M \times K \times K$ columns. Each row contains one of the three-dimensional sliding windows covered by the filters that scan the ifmaps. Filters are reshaped as $N$ column arrays, each consisting of $M \times K \times K$ entries (i.e., the number of weights of each filter). As a result, a GeMM-oriented SA consists of $M \times K \times K$ rows of $N$ PEs each. In addition, $M \times K \times K - 1$ FIFOs must be adopted to ensure proper data-alignment over time. Fig.~\ref{Example_WS}(b) shows an example of Conv-to-GeMM between a $5 \times 5$ ifmap and a $3 \times 3$ kernel. Despite the above PEs' deployment allows WS-based SAs to meet CNNs' workloads, some drawbacks are evident: (i) Conv-to-GeMM introduces data redundancy since inputs are reshaped to guarantee the overlapping sliding windows of the original workflow. As a result, the main memory demands higher capacity, as well as higher number of memory accesses to feed the SA. These aspects negatively impact area and energy efficiency; (ii) furthermore, the larger the kernel size $K$ and/or the number of kernels $M$ and/or the number of filters $N$, the bigger the FIFOs. As a consequence, higher latency and switching activity impact the energy consumption, other than the area.

\subsubsection{Row Stationary SAs}
\begin{figure}
\includegraphics[width=\linewidth]{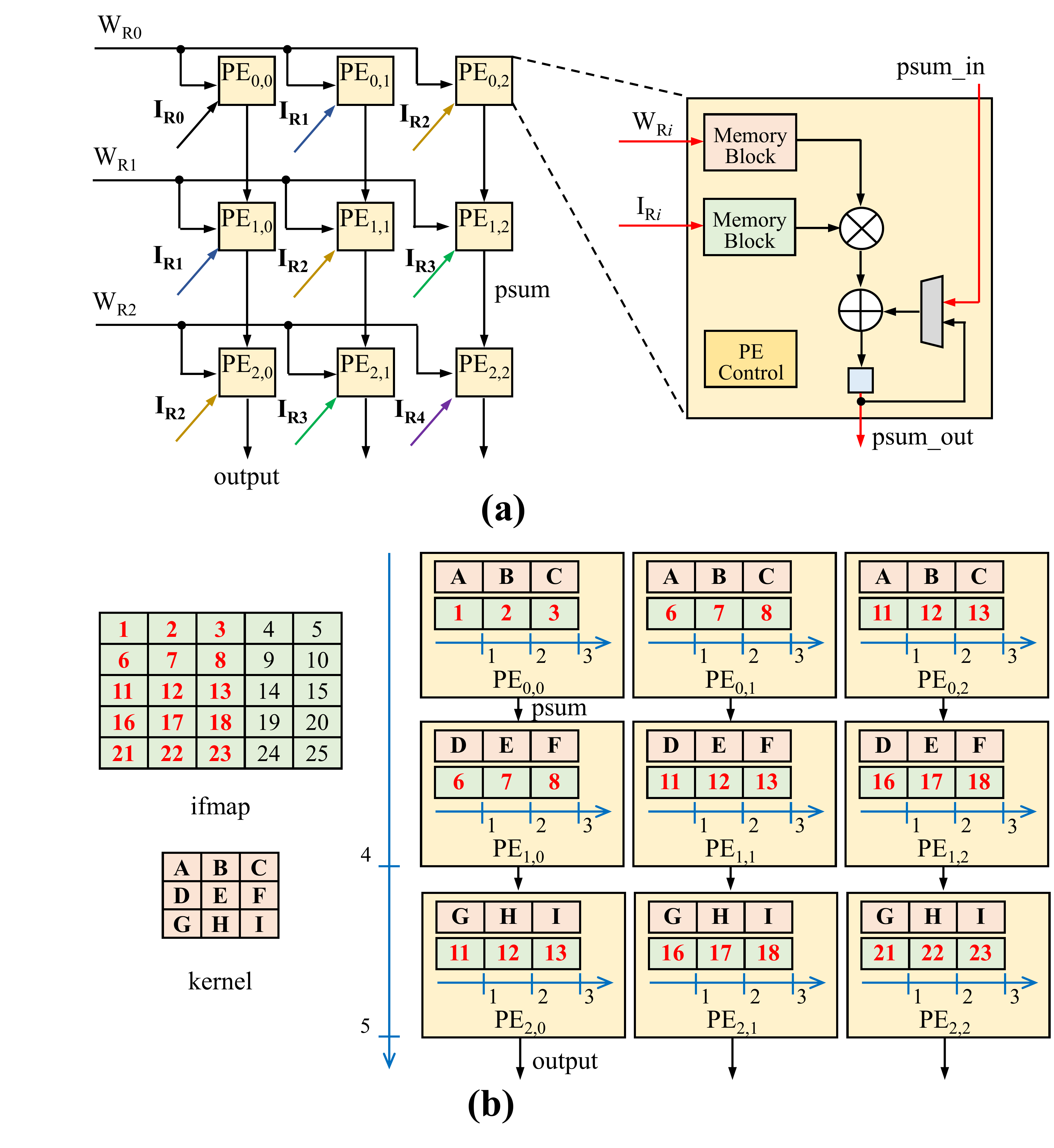}
\centering
\caption{Example of RS-based SA considering a $5 \times 5$ ifmap and a $3 \times 3$ kernel: (a) the architecture consists of $3 \times 3$ PEs, where weights' rows are broadcast horizontally, inputs' rows are broadcast diagonally (in the figure, different numbers and colors refer to different rows). Finally, psums are accumulated vertically. The generic PE, other than accommodating a MAC unit, hosts two memory blocks to circulate rows of inputs and weights, respectively, and a register to save the current psum. In addition, a multiplexer allows the adder to consider either the psum coming from a different PE or the internal psum; (b) an example, where the first three columns of each input row are processed. In each yellow box, which represent a specific PE, the products over time are reported. For instance, the $PE_{0,0}$ executes $1 \times A$ at cycle 1, $2 \times B$ at cycle 2, and $3 \times C$ at cycle 3. For each cycle, the previous product is also accumulated to the current one. At the SA level, psums are accumulated vertically at cycle 4 (first row and second row of the array) and at cycle 5 (second row and third row).}
\label{Example_RS}
\end{figure}
In a RS-based SA, the PEs are arranged as an array of $K$ rows and $H_O$ columns, where: (a) weights are fetched from the main memory and broadcast horizontally to all the PEs; (b) inputs are fetched and broadcast diagonally; (c) psum are accumulated vertically, by exploiting dedicated interconnections between PEs. High data utilization is handled at the PE level, despite requiring memory blocks to circulate inputs and weights. These data are typically loaded as rows of $K$ elements, in order to allow the generic PE to perform 1-D convolutions. In other words, each PE is capable to perform $K$ MACs, before forwarding the temporary psum vertically to another PE, which in the meantime has completed another 1-D convolution. The vertical connections finalize the 2-D convolution. Fig.~\ref{Example_RS}(a) shows an example of RS-based SA using $3 \times 3$ PEs, where a $5 \times 5$ ifmap is subjected to a convolution with a $3 \times 3$ kernel. The generic PE consists of a MAC unit, two memory blocks for inputs and weights, as well as control logic. Fig.~\ref{Example_RS}(b) illustrates the 2-D convolutions of a $5 \times 5$ ifmap, with details on the data shared among the PEs. Differently from Conv-to-GeMM, this dataflow moves data redundancy at the array level, where inputs are shared diagonally, while weights are shared horizontally. In addition, other drawbacks arise: (i) the hardware complexity of each PE is higher, considering that memory blocks (and proper control logic as well) are needed. Other than the area, this negatively impacts the energy dissipation due to the high switching activity of memory blocks to guarantee data circulation; (ii) despite inputs/weights broadcasting at the SA level enables parallel convolutions, thus the possibility to reduce the latency, this directly impacts the routability of the PEs, making the physical implementation phase challenging; (iii) finally, the area covered by the SA depends on ifmap's size, thus limiting scalability and PEs' utilization. 

\begin{figure*}
\includegraphics[width=\textwidth]{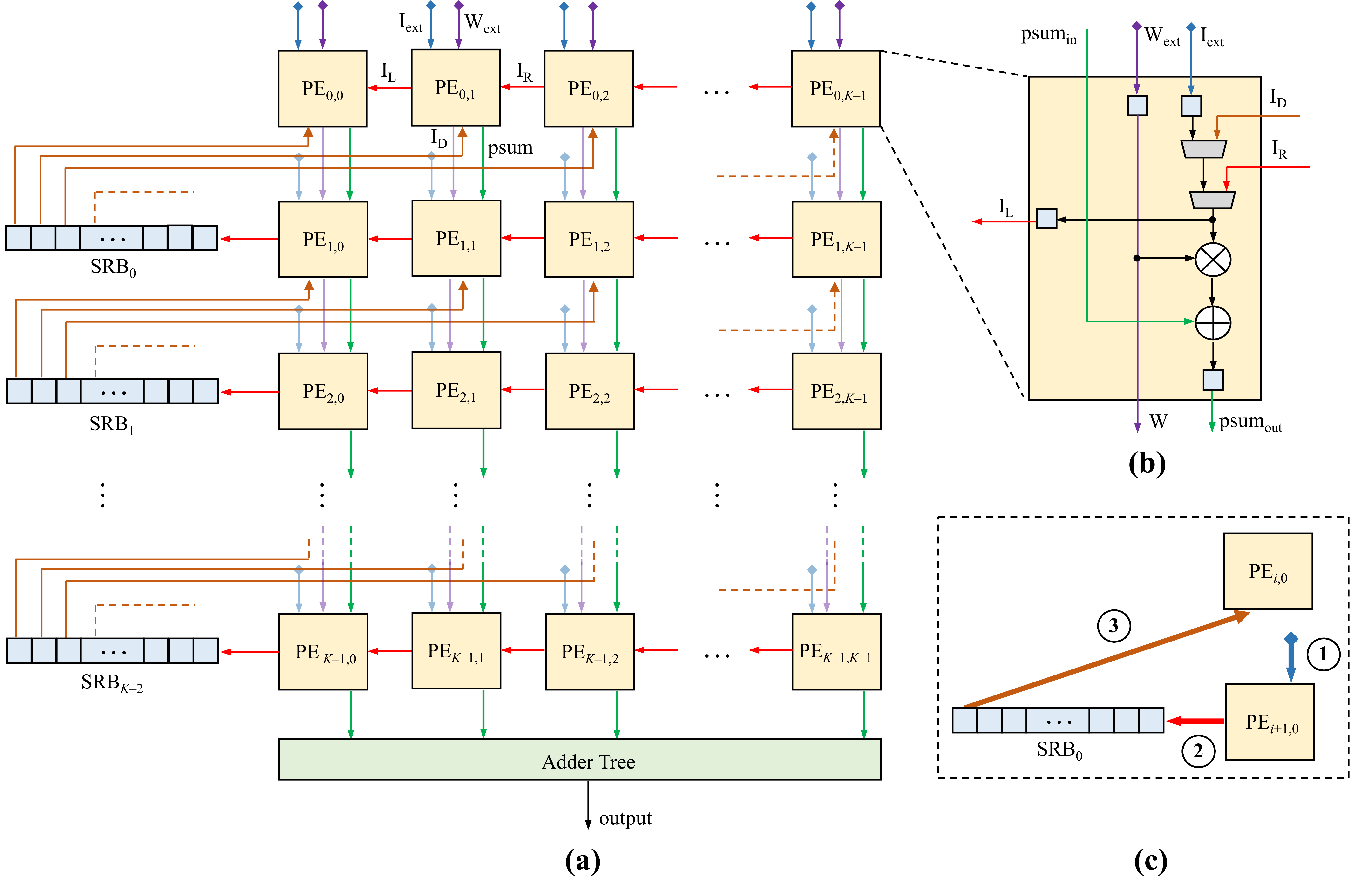}
\centering
\caption{The generic Triangular Input Movement SA. (a) This consists of $K \times K$ PEs, where weights are kept fixed. Inputs are reused internally through either right-to-left movements or bottom to diagonally up translations. Psums are accumulated vertically. Shift Register Buffers (SRBs) assist the diagonal movement. Finally, an adder tree is responsible to accumulate the psums coming from the bottom PEs. (b) The generic PE performs a multiplication between the weight W and the current input, and accumulates this to the psum coming from the top PE ($psum_{in}$), thus generating the current psum ($psum_{out}$). According to a multiplexing logic, the current input may be external ($I_{ext}$), reused from the right side ($I_R$) or reused diagonally from the bottom PE ($I_{D}$). The current input is forwarded to the PE/SRB placed at the left side ($I_{L}$). (c) Example of a triangular input movement between two PEs placed at the left-most side of the array: (1) The generic input is fetched externally and provided vertically to the PE; (2) such input is forwarded from right to left to the SRB; (3) after some cycles, it is supplied diagonally up to the top PE, thus closing the triangle.}
\label{TrIM_generic}
\end{figure*}
\section{Triangular Input Movement Systolic Array}

\begin{figure*}
\includegraphics[width=\textwidth]{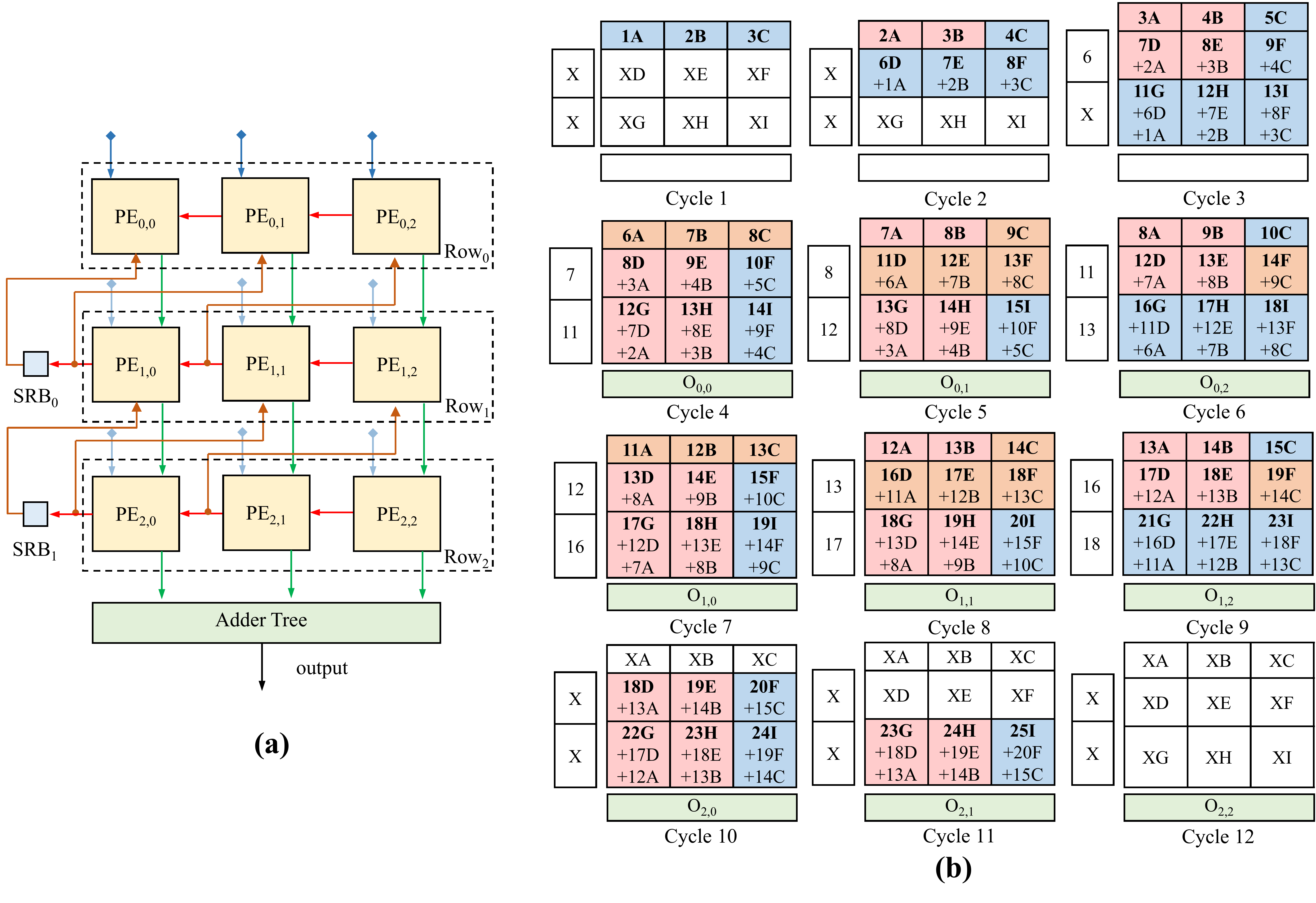}
\centering
\caption{The TrIM architecture and dataflow considering a $5 \times 5$ ifmap and a $3 \times 3$ kernel: (a) the SA consists of $3 \times 3$ PEs and 2 SRBs to enable the diagonal movement of inputs. An extra adder tree, consisting of two adders, accumulates the psums coming from the bottom PEs; (b) the evolution over the different cycles, reporting the computations performed by each PE, the data currently stored into the SRBs and the current output provided by the adder tree. Cells in blue refer to inputs fetched externally, while cells in red indicate inputs reused through a right-to-left movement. Conversely, cells in orange refer to diagonal reuse, either from a SRB or a different PE. Finally, the green boxes refer to the outputs provided by the adder tree. Bold numbers refer to active multiplications. Xs refer to don't care cases. Weights are preloaded through vertical links (not shown in this figure) and keep stationary during all the convolutional computations.}
\label{TrIM_example}
\end{figure*}

\subsection{The Dataflow}
The TrIM-based SA consists of $K \times K$ PEs, where weights are retained at the PE level, while inputs and psums are moved across the array to maximize data reuse. 

Before starting the convolution between the current ifmap and the respective kernel, weights are supplied to the SA. These weights are loaded from top to bottom: in the first cycle, the first group of $K$ weights is supplied to the top row of the array; in the following cycle, the previous group of weights is shifted to the second row, while the first row receives a new group of $K$ weights; and so on, until the entire SA is filled with the $K \times K$ weight kernel. Thus, weights are kept stationary until the completion of all the convolutions.

Inputs supply the PEs through a novel data movement that can be summarized in three steps: (i) first, inputs are fetched from the main memory and delivered to the PEs; (ii) then, such inputs move from right to left, until they reach the left edge of the array; (iii) finally, they move diagonally towards the upper PEs. According to the width of the ifmap under processing, the leftmost PEs may be connected to Shift Register Buffers (SRBs) with depth $W_I - K - 1$, which ensure the correct execution of the diagonal movement over time. Steps (i) to (iii) compose a right-triangular shape, hence the name \textit{Triangular Input Movement} (TrIM). Fig.~\ref{TrIM_generic} schematizes a generic architecture coping with the proposed dataflow. Each PE is labeled as $PE_{i,j}$, with $0 \leq i,j < K$, and consists of a MAC unit, four registers, and two multiplexers to establish whether the current input is reused from a different PE or not. If reused, inputs can move from right to left (thus from each $PE_{i,j+1}$ to $PE_{i,j}$), or they can be forwarded diagonally. In the latter case, they are first provided to the SRBs from each $PE_{i,0}$, then shifted along such buffers, and finally the inputs stored into the last $K$ registers are supplied to the top $PE_{i-1,j}$ elements. In general, when the generic input is grabbed from the memory, the data $I_{ext}$ is supplied to the PE; when the input stored in the right-adjacent PEs is moved into the current PE, thus the data $I_R$ is the target input; finally, when a diagonal movement of the input (either from a bottom PE or from a bottom SRB) has to feed the PE, thus the data $I_D$ is provided. Algorithm~\ref{Algo_TrIM} pitches the orchestration of the three movements over time.

\begin{algorithm}
\fontsize{8.5pt}{9pt}\selectfont
%\color{blue}
\caption{TrIM Dataflow}\label{Algo_TrIM}
% \begin{small}
\begin{algorithmic}[1]
\STATE $\alpha \gets 0$
\FOR{$t=0$ to $H_O\times W_O+K-2$}
    \FOR{$i=0$ to $K-1$}
        \FOR{$j=0$ to $K-1$}
        
            \IF{$i=0$}
                \IF{$t=0$}
                    \STATE $PE_{i,j}(t) \gets I_{ext}$
                \ELSIF{$1 \leq t \leq W_O-1$}
                    \IF {$j=K-1$}
                        \STATE $PE_{i,j}(t) \gets I_{ext}$
                    \ELSE
                        \STATE $PE_{i,j}(t) \gets I_R$
                    \ENDIF
                \ELSIF{$t \bmod W_O = 0$}
                    \STATE $\alpha \gets \alpha+1$
                    \STATE $PE_{i,j}(t) \gets I_D$
                \ELSIF{$t = \alpha W_O+1$}
                    \IF {$0 \leq j < K-1$}
                        \STATE $PE_{i,j}(t) \gets I_R$
                    \ELSE
                        \STATE $PE_{i,j}(t) \gets I_D$
                    \ENDIF
                \ELSIF{$t \geq H_O\times W_O$}
                    \STATE $PE_{i,j}(t) \gets Idle$
                \ELSE
                    \IF {$0 \leq j < K-1$}
                        \STATE $PE_{i,j}(t) \gets I_R$
                    \ELSE
                        \IF{$W_I \leq 2K$}
                            \STATE $PE_{i,j}(t) \gets I_{ext}$
                        \ELSE
                            \IF {$(\alpha+1)W_O-K < t \leq (\alpha+1)W_O-1$}
                                \STATE $PE_{i,j}(t) \gets I_{ext}$
                            \ELSE
                                \STATE $PE_{i,j}(t) \gets I_D$
                            \ENDIF
                        \ENDIF
                    \ENDIF
                \ENDIF
                
            \ELSIF{$i=K-1$}
                \IF{$t<K-1$}
                    \STATE $PE_{i,j}(t) \gets Idle$
                \ELSIF{$t=K-1$ \OR $(t-K+1) \bmod W_O = 0$}
                    \STATE $PE_{i,j}(t) \gets I_{ext}$
                \ELSE
                    \IF{$0 \leq j < K-1$}
                        \STATE $PE_{i,j}(t) \gets I_R$
                    \ELSE
                        \STATE $PE_{i,j}(t) \gets I_{ext}$
                    \ENDIF
                \ENDIF

            \ELSE
                \STATE $PE_{i,j}(t) \gets PE_{0,j}(t-i)$, with $t \geq i$
             
            \ENDIF
        \ENDFOR
    \ENDFOR
\ENDFOR
\end{algorithmic}
% \end{small}
\end{algorithm}

Psums are accumulated vertically, thus from each $PE_{i,j}$ to each $PE_{i+1,j}$. Eventually, an adder tree accumulates the psums coming from the bottom $PE_{K-1,j}$ elements. For very small ifmaps, with $W_I \leq 2K$, the number of registers may be lower than $K$, thus the diagonal connections may also interest some of the PEs. 

To better understand how TrIM works, we consider the case reported in Fig.~\ref{TrIM_example}, where a $5 \times 5$ ifmaps, with $I = 1,2,...,25$ being the inputs, and a $3 \times 3$ kernel, with $W = A,B,...,I$ being the weights, are considered. The equivalent TrIM architecture is made of 3 rows (named $Row_0$, $Row_1$, and $Row_2$), each having 3 PEs. The leftmost PEs belonging to $Row_1$ and $Row_2$ are connected to SRBs with depth 1. A final adder tree, consisting of two adders, manages the psums coming from the PEs of $Row_2$. Preliminarily, 3 cycles are needed to load the $3 \times 3$ weights, arranged in rows of $3$ weights per cycle. Once all the weights are stored, all the convolutions are performed, as detailed in Fig.~\ref{TrIM_example}. For each cycle, the numbers in the PEs indicate the current computation (new multiplications in bold). The different colors refer to the input movement that has led to that multiplication: cells in blue refer to vertical input movement; cells in red refer to right-to-left movements; cells in orange refer to diagonal movements. Input activations temporarily stored in SRBs are reported using white boxes. Finally, the green boxes report the final output from the adder tree. To facilitate the reading of Fig.~\ref{TrIM_example}, we pick the case of cycle 4 as an example: $Row_0$ multiplies $I = 6,7,8$ with $W = A,B,C$, where $I = 6$ is reused diagonally from $SRB_0$, while $I = 7,8$ are reused diagonally from $PE_{1,0}$ and $PE_{1,1}$, respectively. $Row_1$ multiplies the inputs $I = 8,9,10$ with $W = D,E,F$, and accumulates these with the psums coming from $Row_0$. While $I = 8,9$ are reused through right-to-left movements, $I = 10$ is supplied externally. $Row_2$ multiplies the inputs $I = 12,13,14$ with $W = G,H,I$, and accumulates these with the psums coming from $Row_1$. While $I = 12,13$ are reused through right-to-left movements, $I = 14$ is supplied externally. At this cycle, $SRB_0$ and $SRB_1$ receive $I=7$ and $I=11$, respectively. Finally, the psums coming from $PE_{2,0}$, $PE_{2,1}$, and $PE_{2,2}$ are accumulated through the adder ($O_{0,0}$ result).

Overall, the case reported in Fig.~\ref{TrIM_example} allows us to appreciate how TrIM maximizes input utilization. In fact, the number of total memory accesses from the main memory is 29, of which only 4 accesses refer to inputs read more than once. To better spotlight this, let consider the case $I=13$. This input is read once from the main memory (cycle 3) and reused 8 times (from cycle 4 to cycle 9) through right-to-left and diagonal movements. 

\subsection{Support for Fully-Connected Layers}
The TrIM-based SA is tailored for convolution workloads, overcoming the data redundancy issue introduced by Conv-to-GeMM. However, the architecture can be adapted to support conventional FCLs. In this case ifmaps are treated as one-dimensional vectors. In such a scenario, TrIM behaves as a WS-based SA: weights are loaded in the preliminary stage (from top to bottom, through vertical links). Then, new inputs are supplied to the PEs through parallel vertical injection. Horizontal and diagonal movements are not exploited. Psums are vertically accumulated and later reduced through the adder tree at the bottom of the SA. Overall, data are processed in a pipeline fashion without interruptions over time. During FCLs' execution, SRBs are not exploited. However, considering that CNNs exploit hundreds of CLs and few FCLs\cite{8114708}, the overhead of SRBs in FCL is not a disadvantage in general.

\subsection{Differences with WS and RS Dataflows}

To better highlight what makes TrIM different from WS-based and RS-based SAs, we list below the key highlights in terms of different features.
\begin{itemize}
    \item \textbf{Data redundancy}: WS has data redundancy at the main memory level, which adds a significant energy overhead. RS has data redundancy at the array level through memory blocks, adding still significant energy overhead. While RS moves the redundancy penalty from the off-chip to the on-chip level, TrIM overcomes the input redundancy problem and, moreover, maximizes the data utilization by recycling the same input as much as possible. 
    \item \textbf{Weights reuse}: In WS and TrIM, weights are preliminary fetched from the memory and then fixed at the PE level for the entire duration of the convolution (weight-stationary). In RS, weights are \textit{broadcast horizontally} to a row of PEs. Each PE receives a set of $K$ weights that in turn are circulated through a scratch pad. WS and TrIM do not require any scratch pad at the PE level, thus making their micro-architecture simpler with a lower energy footprint.
    \item \textbf{Inputs reuse}: In WS, inputs are grabbed from the memory and forwarded \textit{horizontally} to PEs through FIFOs for proper synchronization. In RS, inputs are \textit{broadcast diagonally} to multiple PEs. Each PE receives a set of $K$ inputs, circulated through scratch pads as long as required. TrIM introduces innovation on the input movement. After being retrieved from the memory and supplied \textit{vertically} to PEs, inputs are reused through either \textit{horizontal} and \textit{diagonal} movements between PEs. The diagonal movements may be assisted by SRBs when the ifmap is fairly larger than the kernel size. Differently from Eyeriss, the diagonal links in TrIM are between PEs or between SRBs and PEs. No broadcasting is thus exploited.
    \item \textbf{Psums propagation}: In all the dataflows (WS, RS, TrIM), psums move vertically from \textit{top} to \textit{bottom}. However, WS requires FIFOs at the bottom for proper output synchronization, while RS makes the PEs' micro-architecture complex: the adder has both a feedback loop (to behave as an accumulator during the processing of a row of $K$ inputs and weights) and a vertical connection for psums accumulation between PEs.
    \item \textbf{Array sizes}: WS and TrIM make use of $K \times K$ PEs. In contrast, RS is a rectangular array with sizes $K \times H_O$, thus exhibiting more limitations regarding the design scalability and area efficiency. In addition, the presence of scratch pads at the PE level makes the RS even bigger. 
\end{itemize}

\section{An Analytical Model for Systolic Arrays}
\label{analytical_model}
In order to highlight the advantages provided by TrIM, we built an analytical model to explore and evaluate the design space of TrIM, searching for the optimum design points that lead to the physical implementation phase.
The aim of this model is twofold: (i) providing insights about memory accesses, throughput, and local buffers/registers; (ii) comparing TrIM to WS and RS dataflows. 
Without loss of generality, all the metrics refer to a convolution between a $H_I \times W_I$ ifmap and a $K \times K$ kernel. However, these can be easily scaled to manage $M$ ifmaps and $N$ filters of $M$ kernels. 

\subsection{Modelling WS and RS Dataflows}
In what follows, memory accesses are to be considered as generalized accesses, without specific reference on the hierarchy level.
The number of memory accesses (MA) of the WS-based SA is dictated by Conv-to-GeMM and reported in equation~(\ref{ma_ws}):
\begin{equation}
\label{ma_ws}
     MA_{WS} = K^2 \times (H_O \times W_O) 
\end{equation}
This conversion results in data redundancy and higher memory capacity, since ifmap's activations are arranged as $H_O \times W_O$ rows and $K^2$ columns.
On the contrary, since no Conv-to-GeMM is required by the RS-based SA, the MA of the main memory are directly related to the ifmap's sizes ($H_I \times W_I$). However, the RS-based SA needs memory blocks at the PE level, usually implemented as SRAM-based scratch pads, which severely affect the energy footprint. Based on the Eyeriss implementation \cite{7738524}, the scratch pads energy is estimated to be from $\alpha= \sim 12.9 \times$ to $\sim 16.5 \times$ higher than the main memory energy. Therefore, the number of MA is summarized by equation~(\ref{ma_rs}):
\begin{equation}
\label{ma_rs}
     MA_{RS} = (1+\alpha) \times (H_I \times W_I)
\end{equation}
The $\alpha$ factor is retrieved from \cite{7738524}, considering the total energy dissipated by the scratch pads on the best and worst case scenarios of the VGG-16 CNN \cite{Simonyan_15} and normalized on the total off-chip memory energy. Regarding the off-chip memory energy, we have considered \cite{Horowitz_14} to determine the energy per single access.

The throughput (T) is expressed as the ratio between the number of operations carried out by the convolution and the total latency required for its completion, thus indicating how many operations are performed per cycle. First, to get the number of operations (OPs), we consider that $K^2$ multiplications and $K^2 - 1$ additions are performed to produce one output, which almost corresponds to $2 \times K^2$ operations. This is repeated $H_O \times W_O$ times to generate as many outputs. Thus, the total number of operations is given by equation~(\ref{ops}):
\begin{equation}
\label{ops}
     OPs = 2 \times K^2 \times H_O \times W_O
\end{equation}
When the WS-based SA is considered, the latency (L) is given by equation~(\ref{lat_ws}):
\begin{equation}
\label{lat_ws}
     L_{WS} = K^2 + H_O \times W_O - 1
\end{equation}
Every PE requires one clock cycle to generate its own psum. Thus, $K^2$ cycles are required to generate one full output. After that, one valid output is provided per cycle. The RS-based SA behaves differently. Every PE requires $K$ cycles to generate the temporary output of a 1-D convolution. After that, $K-1$ extra cycles are needed to accumulate the psums vertically, in order to complete one 2-D convolution. Thus, L is given by equation~(\ref{lat_rs}):
\begin{equation}
\label{lat_rs}
     L_{RS} = W_O \times (2 \times K -1)
\end{equation}
Considering the number of operations and the latency reported above, T is dictated by equations~(\ref{t_ws}) and~(\ref{t_rs}):
\begin{equation}
\label{t_ws}
     T_{WS} = \frac{2 \times K^2 \times H_O \times W_O}{K^2 + H_O \times
     W_O - 1}
\end{equation}
\begin{equation}
\label{t_rs}
     T_{RS} = \frac{2 \times K^2 \times H_O}{2 \times K -1}
\end{equation}
For the sake of fair comparisons, T needs to be normalized by the number of PEs. Since the WS-based SA consists of $K \times K$ PEs, while the RS-based SA is made of $K \times H_O$ PEs, the T per PE (TPE) is given by equations~(\ref{tpe_ws}) and~(\ref{tpe_rs}):
\begin{equation}
\label{tpe_ws}
     TPE_{WS} = \frac{2 \times H_O \times W_O}{K^2 + H_O 
     \times W_O - 1}
\end{equation}
\begin{equation}
\label{tpe_rs}
     TPE_{RS} = \frac{2 \times K}{2 \times K -1}
\end{equation}

To consider local buffers and registers, FIFOs used in the WS-based SA and memory blocks inside the PEs of the RS-based SA are modelled as equivalent sets of registers. The WS-based SA requires 3 registers per PE, as well as $K^2-1$ FIFOs for inputs. The total number of equivalent registers (Reg) is dictated by equation~(\ref{reg_ws}):
\begin{equation}
\label{reg_ws}
     Reg_{WS} = 3 \times K^2 + \frac{K^2 \times (K^2 - 1)}{2}
\end{equation}
Conversely, the PEs inside the RS-based SA adopts two memory blocks, each equivalent to $K$ registers, and a register for the psum. The total number of registers is given by equation~(\ref{reg_rs}):
\begin{equation}
\label{reg_rs}
     Reg_{RS} = (2 \times K + 1) \times K \times H_O
\end{equation}

\subsection{Modelling TrIM Dataflow}
The major advantage offered by TrIM is the substantial reduction of MA if compared to WS and RS dataflow. Indeed, the high data utilization enabled by the triangular input movement maximizes the number of operations per access. Two contributions concur in modelling the MA metric: (i) the initial read of the ifmap, which is given by $H_I \times W_I$ accesses; (ii) a very limited overhead (OV) to retrieve previous inputs, no more locally available. Therefore, MA is given by equation~(\ref{ma_trim}):
\begin{equation}
\label{ma_trim}
    MA_{TrIM} = H_I \times W_I + OV
\end{equation}
where OV is dictated by equation~(\ref{maov_trim}):
\begin{equation}
\label{maov_trim}
\begin{aligned}
    OV =
    \begin{cases}
        (W_I-K-1) \times (K-1) \times (H_I-K) & W_I < 2K \\
        (K-1)^2 \times (H_I-K) & W_I \geq 2K \\
    \end{cases}    
\end{aligned}
\end{equation}

The total latency is given by the number of outputs to be generated, other than an initial latency to fill the pipeline. This initial latency is dictated by the pipeline of the array (three cycles for the PEs, one cycle for the adder tree). Then, one new output is generated per cycle. Thus, the total L is given by equation~(\ref{lat_trim}):
\begin{equation}
\label{lat_trim}
     L_{TrIM} = K + H_O \times W_O
\end{equation}
As a result, T is given by equation~(\ref{t_trim}):
\begin{equation}
\label{t_trim}
     T_{TrIM} = \frac{2 \times K^2 \times H_O \times W_O}{K + H_O \times
     W_O}
\end{equation}
Since TrIM consists of $K \times K$ PEs, the TPE is given by equation~(\ref{tpe_trim}):
\begin{equation}
\label{tpe_trim}
     TPE_{TrIM} = \frac{2 \times H_O \times W_O}{K + H_O \times
     W_O}
\end{equation}

In order to retrieve the total number of registers, we consider that each PE uses four registers, while extra $K-1$ SRBs, each containing $W_I-K-1$ registers, are needed to ensure the diagonal reuse of inputs. In addition, the adder tree uses one register. Equation~(\ref{reg_trim}) summarizes the total number of registers:
\begin{equation}
\label{reg_trim}
     %Reg_{TrIM} = 3 \times K^2 + (K-1) \times W_I + 2
     Reg_{TrIM} = 4 \times K^2 + (K-1) \times (W_I - K -1) + 1
\end{equation}
The component $4 \times K^2$ relates to the registers available at the PE level (4 for each PE). The second block $(K-1) \times (W_I-K-1)$ refers to the registers that model the SRBs. Finally, the extra register equips the adder tree.

\section{Design Space Evaluation}
In this section, TrIM is compared with the WS-based SA and the RS-based SA through a design space evaluation. Memory accesses, throughput, and number of registers are considered, following the equations introduced in Section~\ref{analytical_model}. For each metric, different kernel sizes and ifmap's sizes are considered. $K$ spans the range 3,5,7, extensively used in state-of-the-art CNNs\cite{Targ_16,Szegedy_15}. Square ifmaps ($H_I = W_I = I$) cover the range $I = [16, 32, 64, 128, 256]$. 
\begin{figure}
\includegraphics[width=\linewidth]{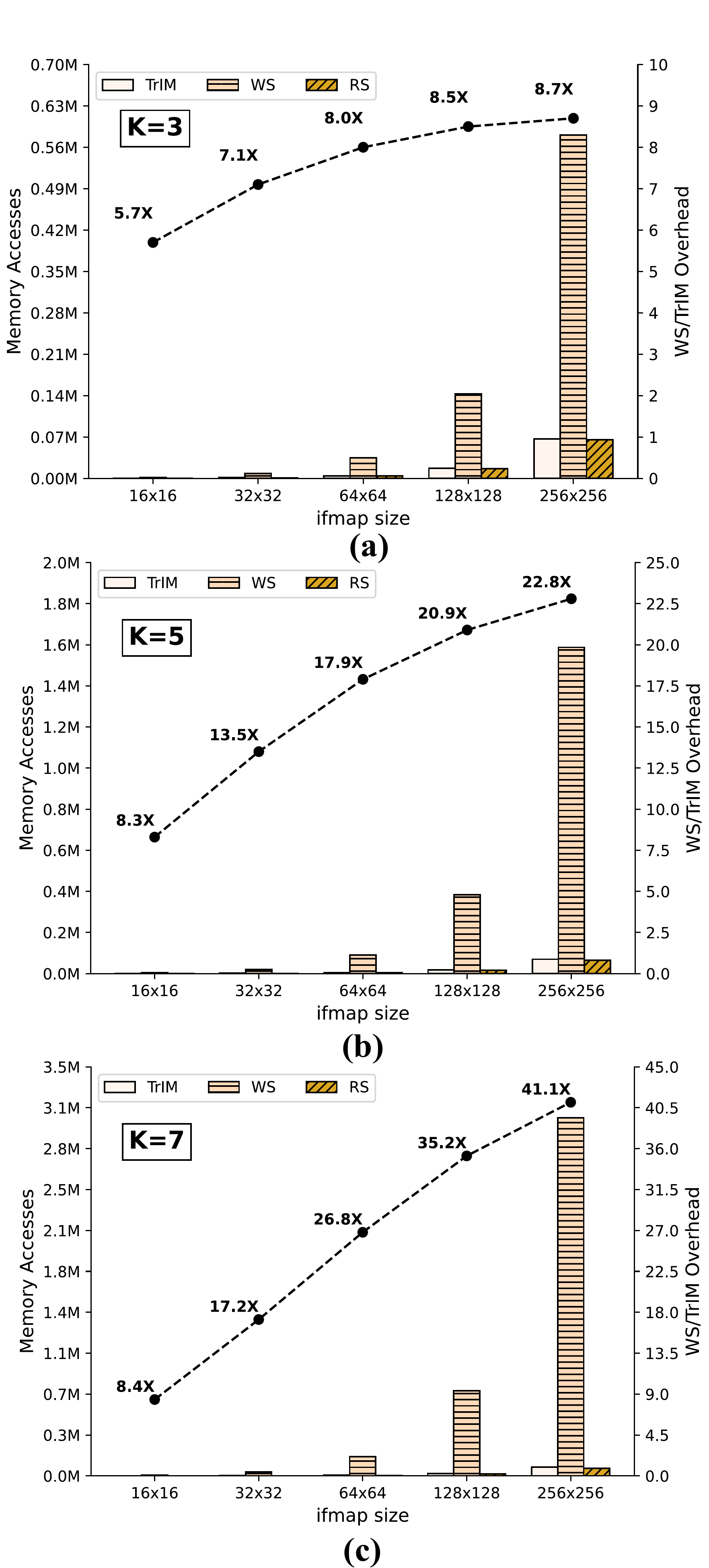}
\centering
\caption{Memory Accesses (MA) analysis. In each plot (a)-(c), the left vertical axis refers to MA, the right vertical axis refers to the ratio between WS and TrIM, whereas the horizontal axis refers to the ifmap size $I$. Bars are associated to the MA metric, dots and lines compare WS and TrIM. With more detail, bars in seashell highlight the MA in TrIM, bars in peachpuff with horizontal hashes are related to WS, while those in goldenrod with diagonal hashes refer to RS. The entire analysis covers $I=16,32,64,128,256$. The bar plots reported in (a) are specific of $K=3$, bar plots in (b) refer to $K=5$, and (c) illustrates the case $K=7$.}
\label{MemAcc}
\end{figure}
\begin{figure}
\includegraphics[width=\linewidth]{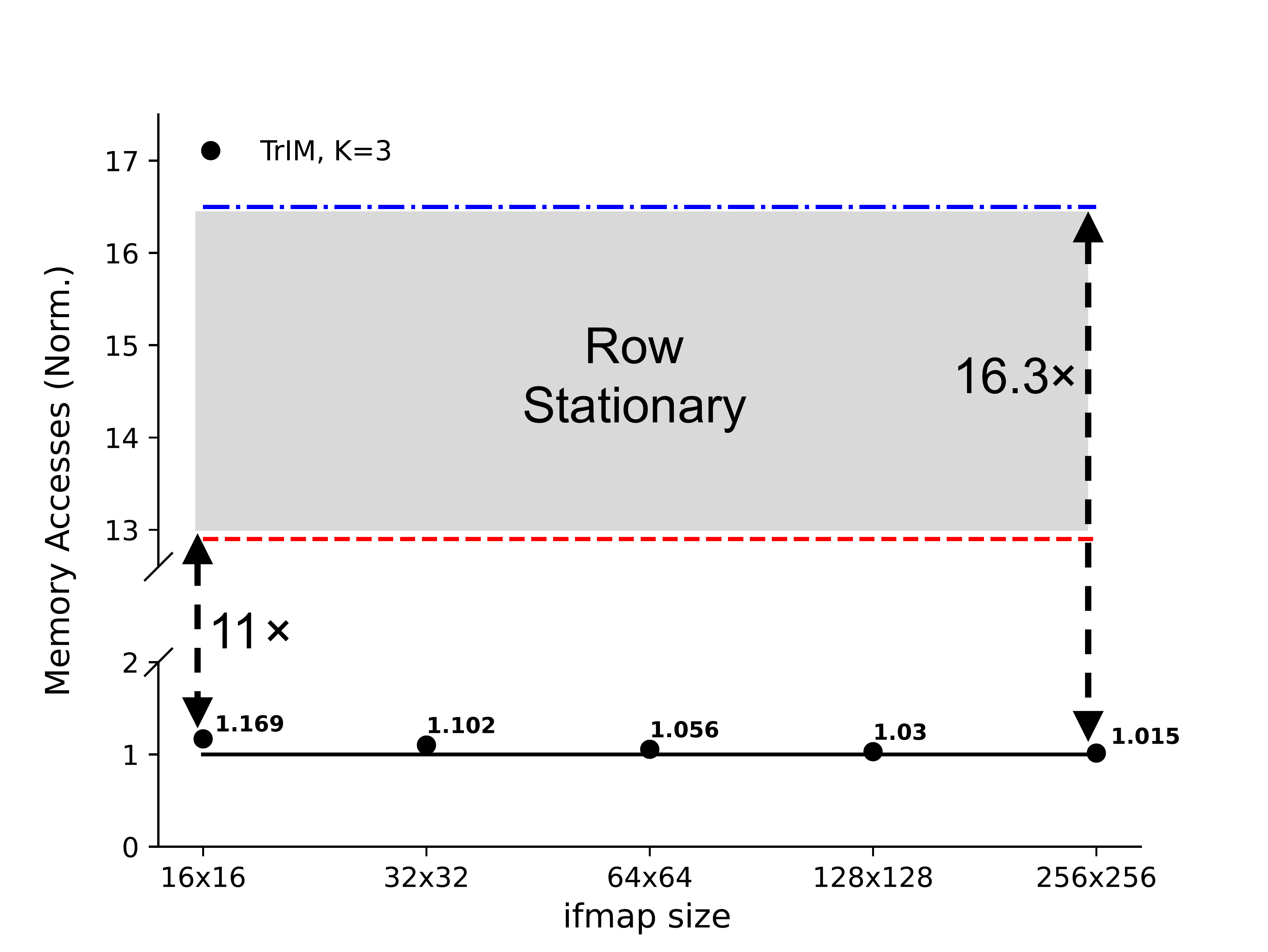}
\centering
\caption{Memory access comparison between TrIM and RS. Memory accesses are normnalized to the number of activations of the generic ifmap, and highlighted by the black line crossing 1.0. Black dots refer to the nromalized memory accesses for TrIM, and varying the ifmap size $I$ from 16 to 256. The gray area between the red and blue dashed lines indicates the spectrum in which the Row Stationary dataflow moves.}
\label{MemAccTrIM_RS}
\end{figure}
\subsection{Memory Accesses}
Fig.~\ref{MemAcc} depicts the number of MA required by the different dataflows. Fig.~\ref{MemAcc}(a) refers to $K=3$, Fig.~\ref{MemAcc}(b) refers to $K=5$, while Fig.~\ref{MemAcc}(c) refers to $K=7$. TrIM significantly outperforms WS since no Conv-to-GeMM is performed, thus not introducing data redundancy and resulting in lower memory capacity. For example, when $K=3$ and $I$ spans from 16 to 256, TrIM requires from $5.7 \times$ to almost one order of magnitude less MA. This improvement is more accentuated with higher $K$, reaching a reduction of up to $41.1 \times$ when $K=7$. 

In the case of the RS dataflow, the total number of MA is given by two contributions: the main memory and the memory blocks for data circulation at the PE level. From the main memory, inputs are read once, therefore the MA corresponds to the number of activations of each ifmap. This contribution is reported in Fig.~\ref{MemAcc}(a)-(c). TrIM's MA are practically close to the RS' main memory accesses. In fact, the overhead introduced by TrIM and previously reported in equation~(\ref{maov_trim}) is minimal. For example, when $K=3$ and $I=256$, TrIM uses only 1.5\% more MA than RS. Fig.~\ref{MemAccTrIM_RS} illustrates the overhead required by RS and dictated by the use of memory blocks at the PE level, usually in the form of scratch pads \cite{7738524} . This contribution is from $\sim 11 \times$ to $\sim 16.3 \times$ higher than the memory accesses required to read data. For example, the factor $11\times$ comes from 12.9/1.169, where $\alpha=12.9$ (Equation~\ref{ma_rs}), while 1.169 relates the normalized memory accesses in TrIM (the 0.169 overhead comes from the analytical model, Equations~\ref{ma_trim} and~\ref{maov_trim}).

\begin{table}[t]
  %\color{blue}
  \centering
   \setlength{\tabcolsep}{4pt} % Default value: 6pt
   \renewcommand{\arraystretch}{0.7} % Default value: 1
  \caption{Normalized Energy Estimation$^{\mathrm{*}}$}
  \begin{tabular}{c c c c c c c c c c}
  \toprule
    \multicolumn{1}{c}{} & \multicolumn{3}{c}{$K=3$} & \multicolumn{3}{c}{$K=5$} & \multicolumn{3}{c}{$K=7$}\\
    \cmidrule(l){2-4} \cmidrule(l){5-7} \cmidrule(l){8-10}
    ifmap size & RS & WS & TrIM & RS & WS & TrIM & RS & WS & TrIM \\
   \midrule
     \\ $16\times16$ & 13.9 & 6.9 & \textbf{1.2} & 13.9 & 14.1 & \textbf{1.7} & 13.9 & 19.1 & \textbf{2.3} \\
     \\ $64\times64$ & 15.7 & 8.4 & \textbf{1.1} & 15.7 & 22.0 & \textbf{1.2} & 15.7 & 40.2 & \textbf{1.5} \\
     \\ $256\times256$ & 17.5 & 8.9 & \textbf{1.0} & 17.5 & 24.2 & \textbf{1.1} & 17.5 & 46.7 & \textbf{1.1} \\
  \bottomrule
   \multicolumn{10}{l}{$^{\mathrm{*}}$Normalized to the energy required to read the generic ifmap once.} \\
  \end{tabular}
  \label{energy_estim}
\end{table}

Table~\ref{energy_estim} reports the normalized energy estimation in the cases $16 \times 16$, $64 \times 64$, and $256\times 256$ ifmaps. The best results are in bold. We refer to the energy needed to read each ifmap once as a baseline, and the numbers are normalized to off-chip memory accesses, following\cite{Horowitz_14}. The contribution of the RS dataflow is mainly affected by the scratch pads' activity at the PE level. In this case, TrIM outperforms RS by up to $17.5\times$. The Conv-to-GeMM conversion influences the higher energy requirements in the WS dataflow. For example, when $K=3$, WS requires from $5.8\times$ to $8.9\times$ more energy than TrIM.

\begin{figure}
\includegraphics[width=0.92\linewidth]{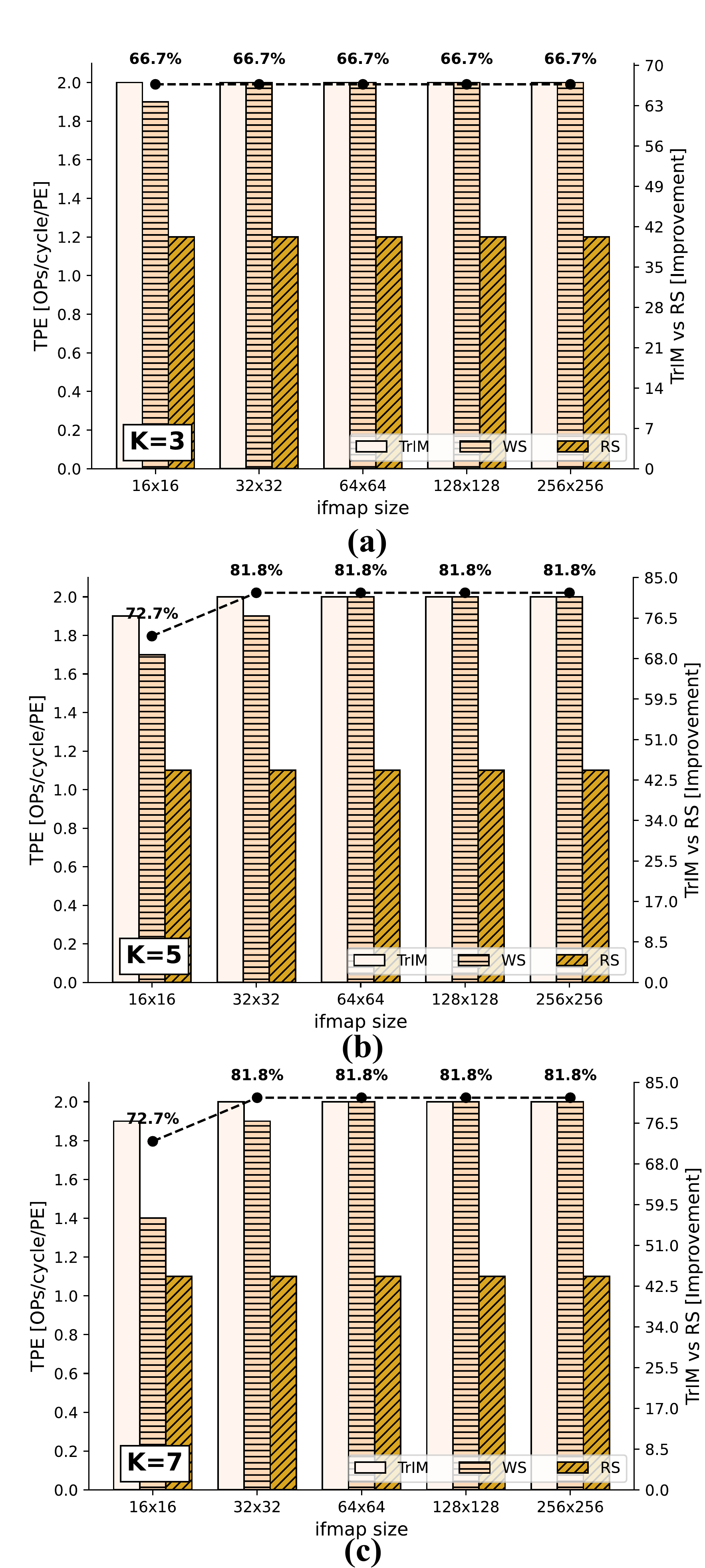}
\centering
\caption{Throughput Analysis: the Throughput per Processing Element (TPE) metric is considered. In each plot (a)-(c), the left vertical axis refers to the TPE, the right vertical axis refers to the percentage improvement between TrIM and RS, whereas the horizontal axis refers to the ifmap size $I$. Bars are associated to the TPE, dots and lines compare TrIM to RS. With more detail, bars in seashell highlight the TrIM performance, bars in peachpuff with horizontal hashes are related to WS, while those in goldenrod with diagonal hashes refer to RS. The entire analysis covers $I=16,32,64,128,256$. The bar plots reported in (a) are specific of $K=3$, bar plots in (b) refer to $K=5$, and (c) illustrates the case $K=7$.}
\label{Throughput}
\end{figure}
\subsection{Throughput}
Fig.~\ref{Throughput} illustrates the TPE for each dataflow. Fig.~\ref{Throughput}(a) refers to $K=3$, Fig.~\ref{Throughput}(b) refers to $K=5$, while Fig.~\ref{Throughput}(c) refers to $K=7$. Considering TrIM, the TPE is independent of $K$ and $I$, and reaches the peak throughput of 2 OPs/cycle/PE. This result spotlights the performance efficiency of PEs coping with the TrIM dataflow. At the PE level, TrIM behaves similarly to the WS-based SA, since weights are not moved. As a result, the TPE of WS and TrIM is the same. However, when small ifmaps are considered, the performance of WS is degraded because of the synchronization requirements of WS. This drawback is even more accentuated with larger $K$. For example, fixed $I=16$, TrIM outperforms WS by $5.3\%$, $11.8\%$, and $35.7\%$ when $K=3,5,7$, respectively.

TrIM also outperforms the TPE of RS. This is because in RS each PE, after processing a row of $K$ inputs and weights, has to wait for almost the same number of cycles to allow psums to be vertically accumulated. Conversely, TrIM is an always-on dataflow, where the accumulations can be overlapped to the multiplications of new inputs and weights. The percentage improvement offered by TrIM is also reported in Fig.~\ref{Throughput}(a)-(c), reaching up to $81.8\%$.

\begin{figure}
\includegraphics[width=\linewidth]{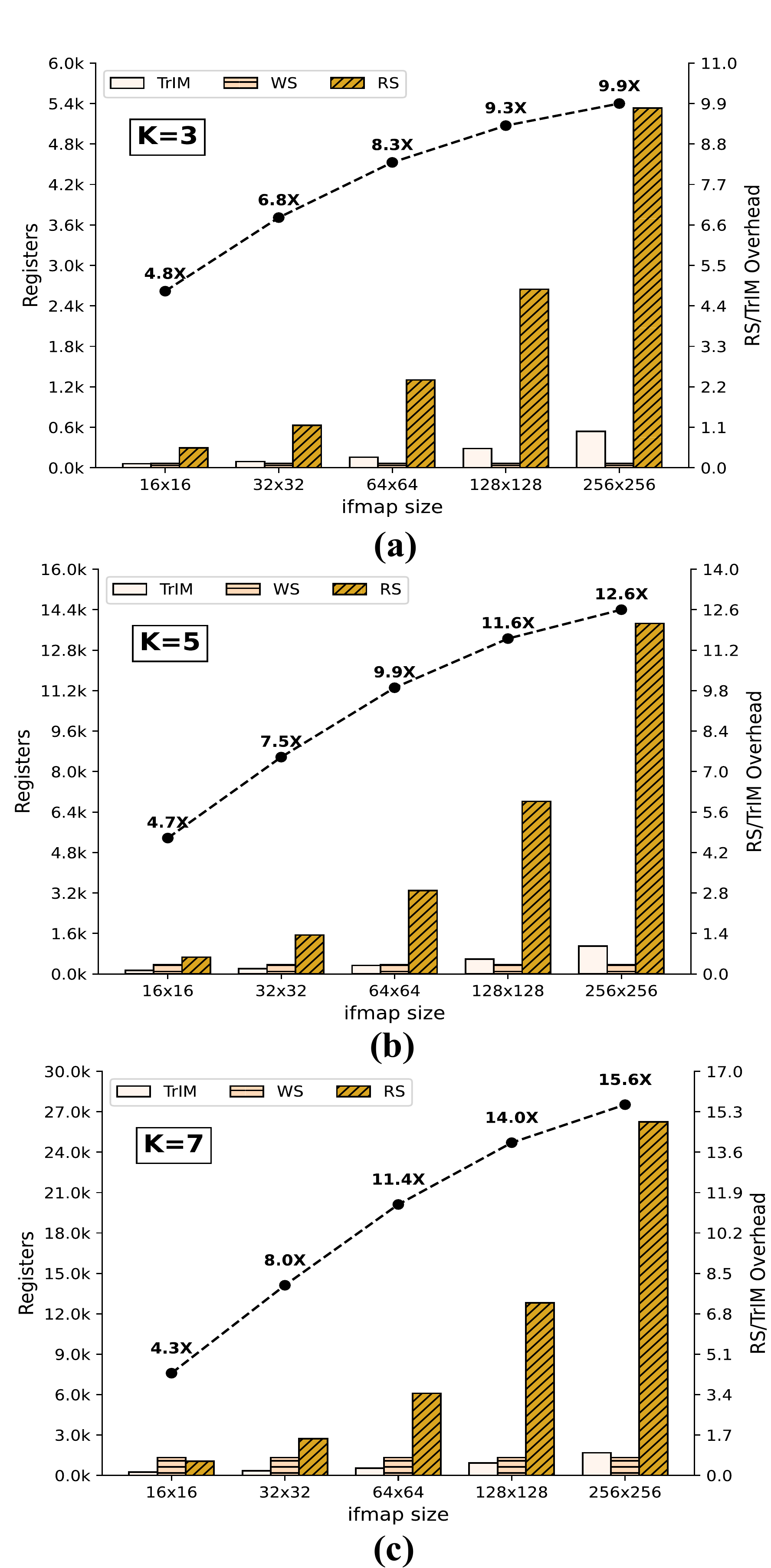}
\centering
\caption{Number of registers: in each plot (a)-(c), the left vertical axis refers to the number of registers, the right vertical axis refers to the overhead between RS and TrIM, whereas the horizontal axis refers to ifmap size $I$. Bars are associated to the number of registers, dots and lines compare RS and TrIM. With more detail, bars in seashell highlight the number of registers in TrIM, bars in peachpuff with horizontal hashes are related to WS, while those in goldenrod with diagonal hashes refer to RS. The entire analysis covers $I=16,32,64,128,256$. The bar plots reported in (a) are specific of $K=3$, bar plots in (b) refer to $K=5$, and (c) illustrates the case $K=7$.}
\label{Registers}
\end{figure}
\subsection{Registers}
Fig.~\ref{Registers} shows the total number of registers for each dataflow. Fig.~\ref{Registers}(a) refers to $K=3$, Fig.~\ref{Registers}(b) refers to $K=5$, while Fig.~\ref{Registers}(c) refers to $K=7$. As explained in Section IV, the number of registers provides the local storage capacity of the different dataflows. Thus, simple registers, FIFOs, scratch pads, and shift registers are all included in the count. The RS-based SA requires the highest number of registers, considering that memory blocks for inputs and weights must be accommodated into $K \times H_O$ PEs. For example, when $K=3$ and $I=256$, the registers of RS are $\sim10\times$ more than those required by TrIM. Since the majority of RS' registers are memory blocks, the degradation in terms of energy is even worse.

When compared to WS, TrIM generally uses more registers if large ifmaps are considered. For example, when $I=64$ and $K=3$, TrIM requires $2.5\times$ more registers. However, the WS-based SA predominantly makes use of FIFOs for data synchronization ($\sim57\%$ in the example case), which are more energy demanding than SRBs in TrIM. 
In contrast, when small ifmaps are considered, TrIM requires fewer registers than WS. This is because SRBs are smaller than FIFOs. In fact, other than dealing with tiny ifmaps, part of the SRBs' workload may be shifted at the PE level (when $I \leq 2K$), where PEs of adjacent rows are diagonally connected. 
The inversion point ($I_{IP}$) between the two trends (i.e., WS better than TrIM and viceversa) follows equation~(\ref{inv_point}), obtained imposing~(\ref{reg_ws}) =~(\ref{reg_trim}), with $W_I=I_{IP}$:
\begin{equation}
\label{inv_point}
     I_{IP} = \Biggl \lceil \frac{K^4-K^2-4}{2 \times(K-1)} \Biggr \rceil
\end{equation}
$I_{IP}=17,75,196$ when $K=3,5,7$, respectively. 

\subsection{Highlights}
The design space analysis presented in this Section allows us to highlight several insights:
\begin{itemize}
    \item TrIM exhibits the lowest number of memory accesses among the competitors. This because of the high data utilization offered by TrIM. In addition, thanks to the triangular movement of inputs, TrIM does not require Conv-to-GeMM, resulting in no data redundancy and, therefore, lower memory capacity than WS. When $K=3$, TrIM outperforms WS by almost $10\times$, and up to $41\times$ with larger $K$. Furthermore, the simple micro-architecture of PEs in TrIM allows significant improvement compared to RS. In fact, RS requires memory blocks at the PE level, in the form of SRAM-based scratch pads, that contribute up to $16.3 \times$ more than the memory accesses associated to the main memory. 
    \item The real throughput achieved by PEs in TrIM matches the peak throughput of 2 OPs/cycle and it is independent by $K$ and $I$. On the contrary, WS exhibits a lower throughput when small ifmaps are considered, due to the FIFO synchronization that negatively impacts the latency. Finally, the computation scheduling in RS significantly degrades the throughput, considering that multipliers must be stalled at regular intervals to allow the vertical accumulations of psums.
    \item TrIM requires from $9.9\times$ to $15.6\times$ fewer registers when compared to RS at different $K$. This is due to the simpler micro-architecture of TrIM's PEs, which do not use memory blocks. Moreover, TrIM is advantageous when also compared to WS and for small ifmaps, considering that SRBs are smaller than synchronization FIFOs in WS. Furthermore, the larger $K$, the larger the ifmaps that can be managed by TrIM using a lower number of registers than WS. This is advantageous with CNNs that make use of different $K$ where, usually, larger $K$ are exploited by larger ifmaps. 

\end{itemize}

\section{Discussion}

This Section provides a perspective on how TrIM can be integrated at the system level for exploitation in hardware architectures for AI. Furthermore, TrIM is contextualized in the current spectrum of dataflows based on systolic arrays, highlighting its unique advantages. Finally, we discuss how TrIM can be exploited in modern Generative AI, such as transformers and diffusion models.

Reconfigurability is one key aspect towards the system-level implementation of TrIM. This is because CNNs consist of diverse CLs, where ifmap and kernel sizes vary. In that perspective, a TrIM-based SA using $K \times K$ PEs can be made adaptable without relevant architectural refactoring. Specifically, the convolution between a $K_E \times K_E$ kernel, with $K_E > K$, and an $H_I \times W_I$ ifmap is realized by splitting the kernel into $K \times K$ tiles, and performing $\lceil K_E/K \rceil \times \lceil K_E/K \rceil$ convolutions, which are later summed together. In the case $(K_E \bmod K) \neq 0$, the kernel tiles may be padded. For example, when $K_E=5$ and $K=3$, four $3 \times 3$ kernel tiles provide the equivalent $5 \times 5$ convolution. System-level management can dynamically pause PEs' activity for padding. From another direction, the TrIM-based SA can support different ifmap sizes by splitting the SRBs into groups. A control logic interfaces each group to the SA, according to the ifmap under processing. A SRB having length $W_{IL}-K-1$, where $W_{IL}$ is the width of the largest ifmap supported by the CNN, may be split into $N_L$ groups, with $N_L$ being the number of CLs having different ifmap sizes. The $K$ registers at the tail of each group are connected, through routing logic, to the PEs.

A special case study is implemented in\cite{Sestito_24_2} to accelerate CNNs like VGG-16\cite{Simonyan_15} and AlexNet\cite{10.1145/3065386}. The architecture, consisting of 168 slices of $3 \times 3$ PEs, has been deployed onto an AMD Zynq UltraScale+ FPGA at a clock frequency of 150 MHz. The results have confirmed the advantage in terms of memory access mitigation, up to $3\times$ less than RS (Eyeriss chip\cite{7738524}). The TrIM architecture exhibits a peak throughput of 453.6 GOPs/s, and completes the VGG-16 inference step within 78.6ms, resulting in $16\times$ speed-up compared to RS. Comparisons with recent FPGA implementations show up to $11.9\times$ higher energy efficiency.

\begin{table*}[t]
    % \color{blue}
    \centering
    \renewcommand{\arraystretch}{0.7} % Default value: 1
    \caption{Comparisons with Enhanced Dataflows}
    \begin{tabular}{c c c c}
    \toprule
    & EOS\cite{8822636} & EWS\cite{10415881} & TrIM \\
    \midrule
    \\ Number of PEs & $G \times P_{H_O} \times P_{W_O}$ & $P_M \times P_N$ & $P_M \times P_N \times K \times K^{\mathrm{*}}$ \\
    \\ Ifmaps Reuse & Inter-PE + ARFs & Inter-PE + ARFs & Inter-PE + SRBs \\
    \\ Weights Reuse & PE Broadcast & WRFs & PE Stationary \\
    \\ Psums Reuse & PE Stationary & Inter-PE + PRFs & Inter-PE \\
    \\ Local Buffers & $G$ ARFs & $P_M$ ARFs + $P_M \times P_N$ WRFs + $P_N$ PRFs & $P_M \times (K-1)$ SRBs \\
    \\ OPs/cycle/PE (peak)$^{\mathrm{**}}$ & $2$ & $2$ & $2$ \\
    \\ Output Bandwidth$^{\mathrm{**}}$ & $G \times P_{H_O} \times P_{W_O} \times B$ & $P_N \times B$ & $P_N \times B$ \\
    \bottomrule
    \multicolumn{4}{l}{$^{\mathrm{*}}$We suppose $P_M \times P_N$ parallel arrays, each having $K \times K$ PEs. $^{\mathrm{**}}$Calculated.} \\
    \end{tabular}
    \label{dataflows_comp}
\end{table*}

\begin{figure}
\includegraphics[width=\linewidth]{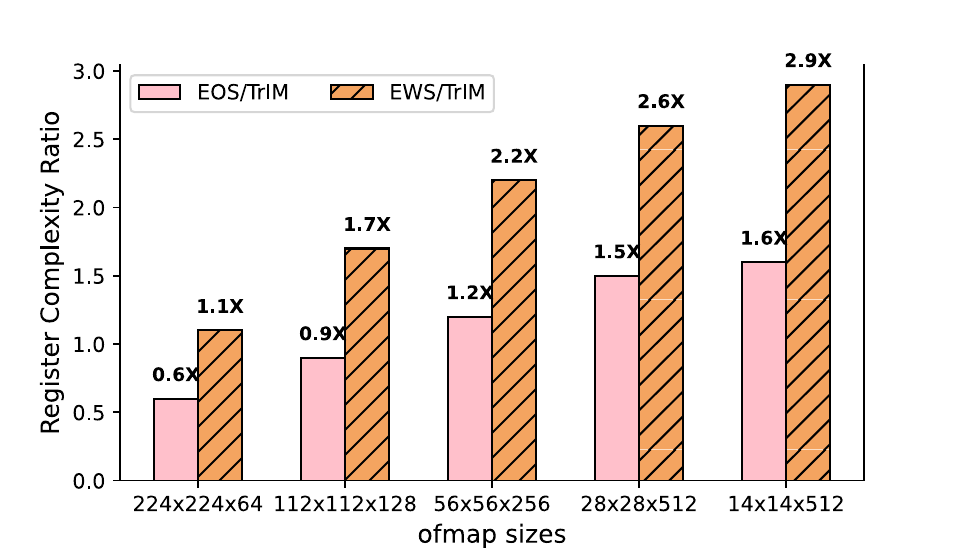}
\centering
\caption{Register complexity ratio between the enhanced dataflows and TrIM. TrIM is considered as the baseline. The analysis refers to 64 PEs. Reuse and parallelism parameters for the enhanced dataflows follow the optimal values as in \cite{8822636,10415881}. Bars in pink are related to the ratio between EOS and TrIM. Dashed bars in orange relate to the ratio between EWS and TrIM. The analysis covers typical ofmap sizes as in the VGG-16 CNN\cite{Simonyan_15} with $K=3$.}
\label{Enhanced_Dataflows_Comp}
\end{figure}

To further steer the uniqueness of TrIM in the SAs' spectrum, we compare its dataflow to the recent literature, where optimizations to baseline dataflows are often introduced. Relevant examples are the Enhanced Output Stationary (EOS)\cite{8822636} and the Enhanced Weight Stationary (EWS)\cite{10415881}.
Table~\ref{dataflows_comp} summarizes the features of these dataflows compared to TrIM. As an example workload, we consider a CL with parameters $H_O$, $W_O$, $M$, $N$ and $K$ as introduced in Section II-A. The parameters $P_{H_O}$, $P_{W_O}$, $P_M$ and $P_N$ relate to the offered parallelism. $B$ is the output bit-width. EOS\cite{8822636} adopts an array split into $G$ groups of $P_{H_O} \times P_{W_O}$ PEs and proposes Activation Register Files (ARFs) to improve ifmap reuse. These ARFs store redundant activations between consecutive sliding windows. EWS\cite{10415881} uses $P_M \times P_N$ PEs with ARFs for redundant activations. In addition, EWS proposes Weight Register Files (WRFs) at the PE level and Psum Register Files (PRFs) at the array level to maximize the reuse of weights and psums. Figure~\ref{Enhanced_Dataflows_Comp} shows the register complexity ratio between EOS, EWS and TrIM. The ratio accounts for ARFs, WRFs, SRBs and PE registers for activations and weights. The analysis is based on 64 PEs. In the case of EOS/TrIM (EWS/TrIM), the register complexity ratio scales from $\sim0.6\times$ ($\sim1.1\times$) to $\sim1.6\times$ ($\sim2.9\times$) when example layers from the VGG-16 CNN\cite{Simonyan_15} are considered. All the compared SAs exhibit a peak throughput of 2 OPs/cycle/PE. However, the real performance is affected by the required bandwidth. For instance, Table~\ref{dataflows_comp} reports the analytical expression of the output bandwidth. It can be seen that EOS delivers $G \times P_{H_O} \times P_{W_O}$ parallel convolutions, which may constraint the real throughput and the scalability of the array.

Other works combine the capabilities of the different dataflows towards hybrid architectures. ReDas\cite{10527420} supports multiple dataflows for DNN benchmarking: complex PEs are used (with data supplied from four directions) with the support of multi-mode buffers at the array edges. TrIM, which is tailored for convolutions, simplifies the complexity of PE and buffer requirements, but preserves high data utilization: weights are reused at the PE level, ifmap are reused through triangular movement at the array level.

The effectiveness of TrIM to deal with convolutions can be also exploited in modern Generative AI: for instance, the Convolutional Vision Transformer (CvT)\cite{Wu_2021_ICCV} adopts convolutional token embedding layers and convolutional projection layers to combine the invariance properties of convolutions with the global generalization and dynamic attention of transformers. Furthermore, Stable Diffusion models benefit from convolutional layers in text-to-image translation. For example, ControlNet\cite{Zhang_2023_ICCV} makes use of U-Net\cite{10.1007/978-3-319-24574-4_28}, which is an encoder-decoder architecture based on convolutions for down-sampling and up-sampling purposes.

\section{Conclusion}
This paper introduces TrIM, an innovative dataflow for SAs that is compatible with CNN computing. The dataflow has two levels of granularity: at the PE level, weights are kept stationary; high input utilization is guaranteed at the SA level, based on a triangular movement. To achieve this, $K \times K$ PEs are interconnected with each other in different directions, and local shift registers are placed on the left edge of the array to assist the triangular movement of inputs. 
We built an analytical model to characterize the array before the physical implementation phase. Memory accesses, throughput, and number of registers are subjected to in-depth design space exploration. When compared to WS dataflow, TrIM requires one order of magnitude fewer memory accesses considering that no data redundancy is exploited. Furthermore, TrIM requires less memory access than RS since no micro-architectural memory blocks assist data circulation. The simpler PEs in TrIM also result in $15.6\times$ fewer registers than RS. Finally, thanks to the overlap between multiplications and accumulations, TrIM's real throughput achieves the peak throughput of 2 OPs/cycle/PE, overcoming RS by 81.8\%.  

\bibliographystyle{IEEEtran}
\bibliography{IEEEabrv,bib}

\begin{IEEEbiography}
[{\includegraphics[width=1in,height=1.25in,clip,keepaspectratio]{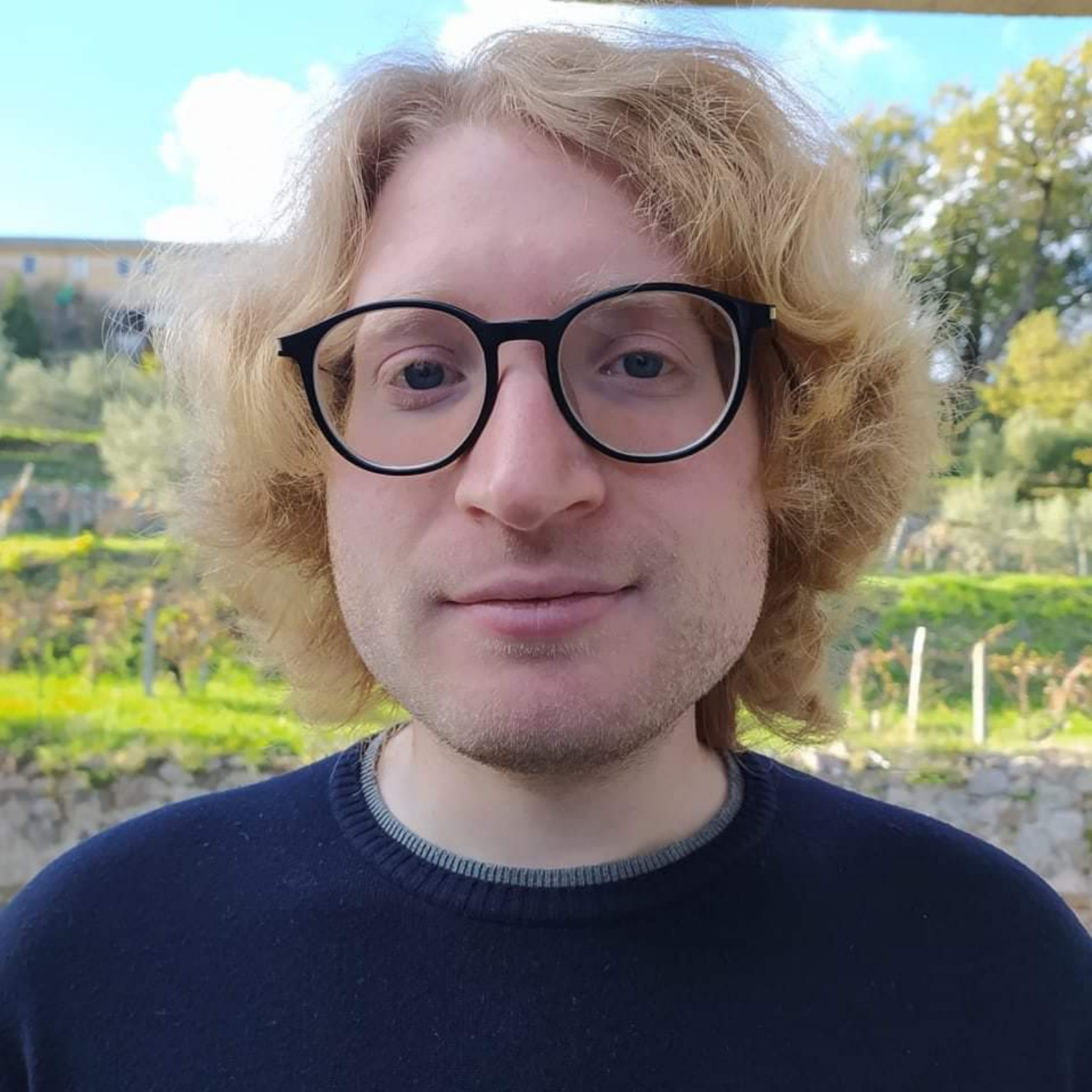}}]{Cristian Sestito}
(Member, IEEE) is a Research Fellow at the Centre for Electronics Frontiers CEF, The University of Edinburgh (UK). He received his BSc and MSc degree from University of Calabria (Italy), both in Electronic Engineering. He got his PhD in Information and Communication Technologies from the same university in 2023, focusing on Convolutional Neural Networks and their implementation on Field Programmable Gate Arrays (FPGA). In 2021/2022, Cristian was a Visiting Scholar at Heriot-Watt University, Edinburgh, working on neural networks’ compression. His research interests include digital design, embedded system design for AI on FPGA-based systems-on-chip, software simulators for neuromorphic AI, AI for circuits and systems design automation.
\end{IEEEbiography}

\begin{IEEEbiography}
[{\includegraphics[width=1in,height=1.25in,clip,keepaspectratio]{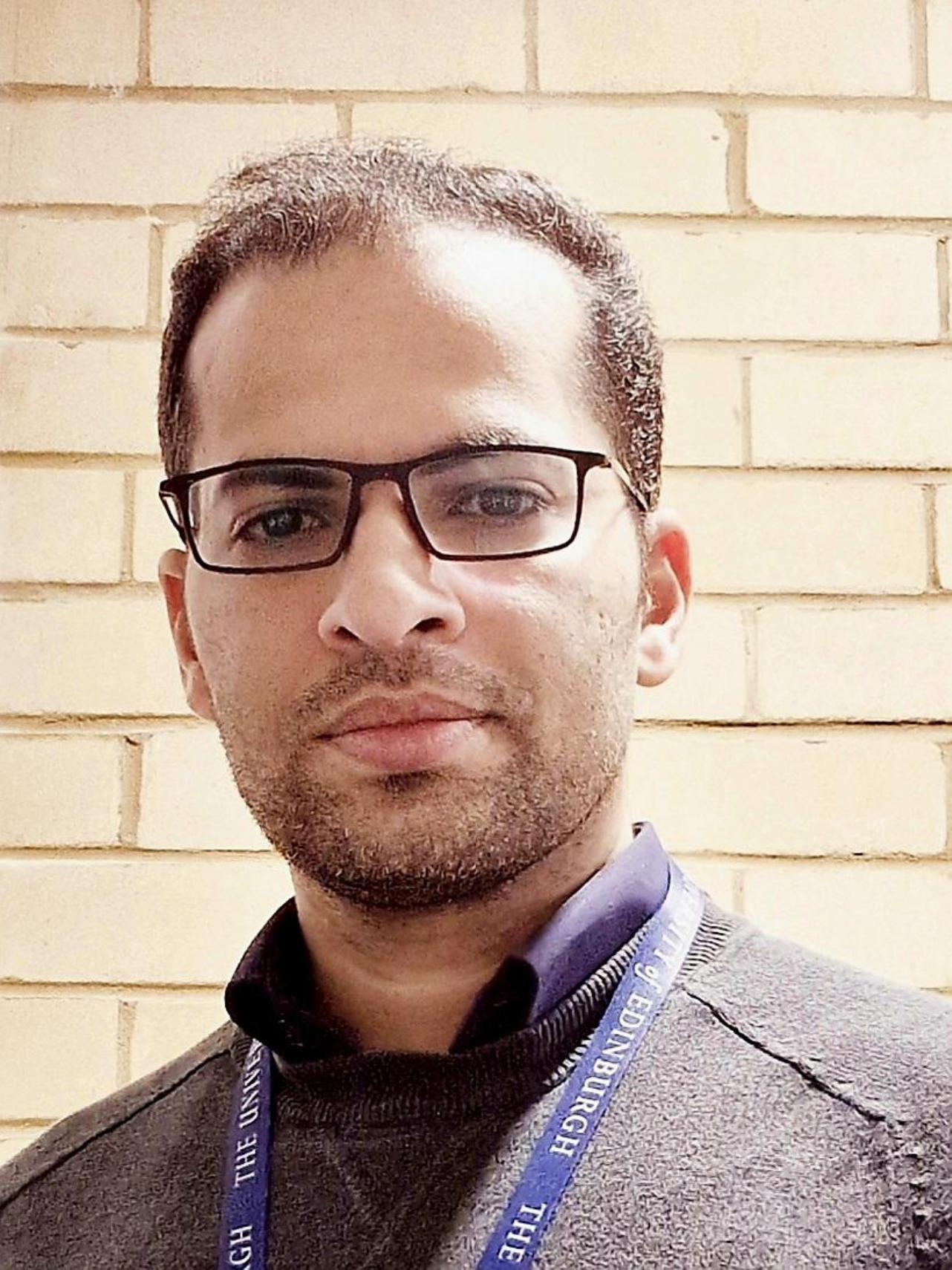}}]{Shady Agwa}
(Member, IEEE) is a Research Fellow at the Centre for Electronics Frontiers CEF, The University of Edinburgh (UK). He received his BSc and MSc degree from Assiut University (Egypt), both in Electrical Engineering. He got his PhD in Electronics Engineering from The American University in Cairo (Egypt) in 2018. Following his PhD, he joined the Computer Systems Laboratory at Cornell University (USA) as a Postdoctoral Associate for two years. In 2021, Shady joined the Centre for Electronics Frontiers at the University of Southampton (UK) as a Senior Research Fellow and then as a Research Fellow at the University of Edinburgh (UK). His research interests span across VLSI and Computer Architecture for AI using conventional and emerging technologies. His work focuses on ASIC-Driven AI Architectures with extensive expertise in In-Memory Computing, Stochastic Computing, Systolic Arrays, Beyond Von Neumann Architectures, Memories and Energy-Efficient Digital ASIC Design.
\end{IEEEbiography}

\begin{IEEEbiography}
[{\includegraphics[width=1in,height=1.25in,clip,keepaspectratio]{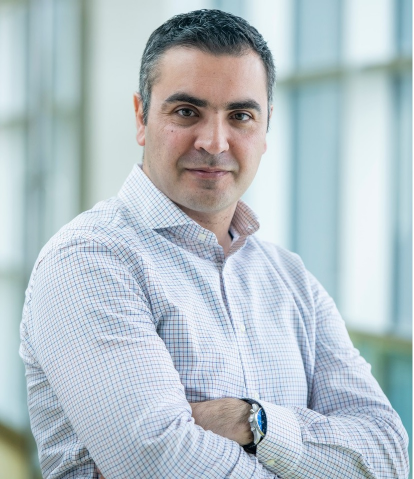}}]{Themis Prodromakis}
(Senior Member, IEEE) received the bachelor’s degree in electrical and electronic engineering from the University of Lincoln, U.K., the M.Sc. degree in microelectronics and telecommunications from the University of Liverpool, U.K., and the Ph.D. degree in electrical and electronic engineering from Imperial College London, U.K. He then held a Corrigan Fellowship in nanoscale technology and science with the Centre for Bio-Inspired Technology, Imperial College London, and a Lindemann Trust Visiting Fellowship with the Department of Electrical Engineering and Computer Sciences, University of California at Berkeley, USA. He was a Professor of nanotechnology at the University of Southampton, U.K. He holds the Regius Chair of Engineering at the University of Edinburgh and is Director of the Centre for Electronics Frontiers. He is currently a Royal Academy of Engineering Chair in emerging technologies and a Royal Society Industry Fellowship. His background is in electron devices and nanofabrication techniques. His current research interests include memristive technologies for advanced computing architectures and biomedical applications. He is a fellow of the Royal Society of Chemistry, the British Computer Society, the IET, and the Institute of Physics.
\end{IEEEbiography}

\end{document}